\documentclass[acmsmall]{acmart}
\makeatletter
\let\@authorsaddresses\@empty
\makeatother

\usepackage[utf8]{inputenc} 
\usepackage[T1]{fontenc}    
\usepackage{booktabs}       
\usepackage{amsfonts}       
\usepackage{nicefrac}       
\usepackage{microtype}      
\usepackage{amsmath,amsthm} 

\usepackage{amssymb} 

\usepackage{mathtools}
\usepackage{color}
\usepackage[font=small,labelfont=bf]{caption}
\usepackage{subfigure}
\usepackage{relsize}
\usepackage{enumitem}
\usepackage{multirow,multicol}

\theoremstyle{definition}
\newtheorem{definition}{Definition}[section]
\theoremstyle{remark}

\usepackage{etoolbox,siunitx}
\usepackage{tabularx}
\usepackage{array}
\usepackage{hyperref}

\usepackage{makecell}

\DeclareMathOperator{\A}{\mathbf{A}}
\DeclareMathOperator{\An}{\Tilde{\mathbf{A}}}

\DeclareMathOperator{\V}{\mathbf{V}}
\DeclareMathOperator{\E}{\mathbf{E}}

\DeclareMathOperator{\Ll}{\mathbf{L}}
\DeclareMathOperator{\Ln}{\Tilde{\mathbf{L}}}

\DeclareMathOperator{\D}{\mathbf{D}}
\DeclareMathOperator{\Dn}{\Tilde{\mathbf{D}}}
\DeclareMathOperator{\Drw}{\mathbf{D}^{-1}}
\DeclareMathOperator{\Dsym}{\mathbf{D}^{-\frac{1}{2}}}

\DeclareMathOperator{\Dnsym}{\Tilde{\mathbf{D}}^{-\frac{1}{2}}}

\DeclareMathOperator{\Z}{\mathbf{Z}}
\DeclareMathOperator{\X}{\mathbf{X}}
\DeclareMathOperator{\h}{\mathbf{h}}

\DeclareMathOperator{\N}{\mathcal{N}}

\DeclareMathOperator{\Pp}{\mathbf{P}}
\DeclareMathOperator{\Qq}{\mathbf{Q}}
\DeclareMathOperator{\I}{\mathbf{I}}

\DeclarePairedDelimiter{\diagfences}{(}{)}
\newcommand{\diag}{\operatorname{diag}\diagfences}

\DeclareMathOperator{\UU}{\mathbf{U}}
\DeclareMathOperator{\UT}{\mathbf{U^{\intercal}}}
\DeclareMathOperator{\uu}{\mathbf{u}}
\DeclareMathOperator{\ut}{\mathbf{u^{\intercal}}}
\DeclareMathOperator{\g}{\mathbf{g}}
\DeclareMathOperator{\LBD}{\mathbf{\Lambda}}

\newcommand\ChangeRT[1]{\noalign{\hrule height #1}}

\usepackage{wasysym}

\renewcommand\footnotetextcopyrightpermission[1]{} 

\AtBeginDocument{%
  \providecommand\BibTeX{{%
    \normalfont B\kern-0.5em{\scshape i\kern-0.25em b}\kern-0.8em\TeX}}}
    
\title{Bridging the Gap between Spatial and Spectral Domains: A Unified Framework for Graph Neural Networks}
\author{Zhiqian Chen}
\email{zchen@cse.msstate.edu}
\affiliation{%
  \institution{of Computer Science and Engineering, Mississippi State University}
  \streetaddress{Butler Hall}
  \city{MS State}
  \state{Mississippi}
  \postcode{39762}
  \country{U.S.A}
}

\author{Fanglan Chen}
\email{fanglanc@vt.edu}
\affiliation{%
  \institution{Computer Science, Virginia Tech}
  \streetaddress{7054 Haycock Road}
  \city{Falls Church}
  \state{Virginia}
  \postcode{22043}
  \country{U.S.A}
}

\author{Lei Zhang}
\email{zhanglei@vt.edu}
\affiliation{%
  \institution{Computer Science, Virginia Tech}
  \streetaddress{7054 Haycock Road}
  \city{Falls Church}
  \state{Virginia}
  \postcode{22043}
  \country{U.S.A}
}

\author{Taoran Ji}
\email{jtr@vt.edu}
\affiliation{%
  \institution{Computer Science, Virginia Tech}
  \streetaddress{7054 Haycock Road}
  \city{Falls Church}
  \state{Virginia}
  \postcode{22043}
  \country{U.S.A}
}

\author{Kaiqun Fu}
\email{fukaiqun@vt.edu}
\affiliation{%
  \institution{Electrical Engineering \& Computer Science, South Dakota State University}
  \country{U.S.A}
}

\author{Liang Zhao}
\email{liang.zhao@emory.edu}
\affiliation{%
  \institution{Computer Science, Emory University}
  \city{Fairfax}
  \country{U.S.A}
}

\author{Feng Chen}
\email{feng.chen@utdallas.edu}
\affiliation{%
 \institution{Computer Science, The University of Texas at Dallas}
 \streetaddress{ECSS 3.901 UTD}
 \city{Dallas}
 \state{Texas}
 \country{U.S.A}}

\author{Lingfei Wu}
\email{lingfei.wu@jd.com}
\affiliation{%
 \institution{JD.COM Silicon Valley Research Center}
 \city{Mountain View}
 \state{California}
 \postcode{94043}
 \country{U.S.A}}

\author{Charu Aggarwal}
\email{charu@us.ibm.com}
\affiliation{%
 \institution{IBM T. J. Watson Research Center}
 \country{U.S.A}}

\author{Chang-Tien Lu}
\affiliation{%
  \institution{Computer Science, Virginia Tech}
  \streetaddress{7054 Haycock Road, Room 312}
  \city{Falls Church}
  \state{Virginia}
  \postcode{22043}
  \country{U.S.A}
}
\email{ctlu@vt.edu}

\begin{document}



\begin{abstract}
Deep learning's performance has been extensively recognized recently. 
Graph neural networks (GNNs) are designed to deal with graph-structural data that classical deep learning does not easily manage.
Since most GNNs were created using distinct theories, direct comparisons are impossible. Prior research has primarily concentrated on categorizing existing models, with little attention paid to their intrinsic connections. The purpose of this study is to establish a unified framework that integrates GNNs based on spectral graph and approximation theory. The framework incorporates a strong integration between spatial- and spectral-based GNNs while tightly associating approaches that exist within each respective domain.
\end{abstract}

\maketitle
\pagestyle{plain}

\section{Introduction}

Deep learning's performance in various machine learning tasks \cite{lecun2015deep,redmon2016you,ren2015faster,hinton2012deep,wu2016google,luong2015effective} has been extensively recognized in recent decades, with amazing success on Euclidean data. 
In recent decades, a slew of new applications have emerged in which effective information analysis boils down to the non-Euclidean geometry of data represented by a graph, such as social networks \cite{lazer2009life}, transportation networks \cite{bell1997transportation}, spread of epidemic disease \cite{newman2002spread}, brain's neuronal networks \cite{marx2012high}, gene data on biological regulatory networks \cite{davidson2002genomic}, telecommunication networks \cite{drew2008diagnosing}, and knowledge graph \cite{lin2015learning}. 
Previous deep learning algorithms, such as convolutional and recurrent neural networks, couldn't handle such non-Euclidean problems on graph-structured data.
Modeling data using a graph is difficult because graph data is irregular, i.e., each graph has a different number of nodes and each node in a graph has a varied number of neighbors, making some operations like convolutions inapplicable to the network structure.

There has recently been a surge growing interest in applying deep learning to graph data. Inspired by deep learning's success, principles from deep learning models are used to handle the graph's inherent complexity. This growing trend has piqued the interest of the machine learning community, and a huge number of graph neural networks (GNN) models have been proposed based on diverse theories \cite{bruna2014spectral,kipf2016semi,defferrard2016convolutional,hamilton2017inductive,atwood2016diffusion,velivckovic2017graph} and grouped into coarse-grained groupings like the spectral \cite{monti2017geometric,hamilton2017representation,zhang2018deep,zhou2018graph,wu2019comprehensive} and spatial \cite{hamilton2017inductive,atwood2016diffusion,velivckovic2017graph} domains.
GNNs have seen rising popularity recently for learning graph representations and quickly spread out to many application domains, such as 
physics \cite{alet2019graph,kipf2018neural}, 
chemistry \cite{dai2019retrosynthesis,bradshaw2018generative}, 
knowledge graph \cite{arora2020survey,ye2022comprehensive,zhang2020efficient,xu2019dynamically}, 
recommender systems \cite{wu2020graph,chen2020revisiting,fan2019graph,wang2021graph}, 
computer vision \cite{jiao2022graph,gao2019know,li2017situation}, 
natural language processing \cite{wu2021graph,sorokin2018modeling,beck2018graph}, 
combinatorial optimization \cite{sato2019approximation,cappart2021combinatorial,gasse2019exact}, traffic network \cite{jiang2022graph,bui2021spatial,ye2020build,chen2020graph}, 
program representation \cite{dinella2020hoppity,wei2020lambdanet,zhou2019devign}, 
social network \cite{bian2020rumor,wu2020graph,peng2019fine}.
However, current research in methodology has not translated into a clear understanding of the mechanisms involved, nor has it given us insight into GNNs' effectiveness or physical meaning. 
As a result, several consequences will occur: 
(1) There is no underlying principle that connects all GNNs, which also limits their growth. 
(2) In high-stakes applications such as drug development, GNN models may carry potentially hazardous unknowns since they are black boxes.
Consequently, the necessity of dissecting GNNs is highlighted, thereby driving academics to hunt for a more universal framework.
The main problem is that existing GNN models use a variety of techniques, including random walks \cite{perozzi2014deepwalk,grover2016node2vec,feng2020graph}, PageRank \cite{bojchevski2020scaling,chien2020adaptive,klicpera2018predict,klicpera2018combining}, attention models \cite{velivckovic2017graph,knyazev2019understanding,kim2020find}, low-pass filters \cite{nt2019revisiting,Bo2020beyond}, and message forwarding \cite{gilmer2017neural,geerts2021let}.
Some preliminary research can only explain a few GNNs methods \cite{xu2018how,gilmer2017neural,ying2019gnn}, leaving the majority of GNN unaccounted for. 
Previous GNNs surveys have dealt mostly with classifying several existing models into multiple categories and expanding on each category individually, with no regard to the interrelationships between them \cite{monti2017geometric,hamilton2017representation,zhang2018deep,zhou2018graph,wu2019comprehensive}.


This research \footnote{A short version is available at \cite{chen2020bridging}} aims to provide a coherent framework for generalizing GNNs by bridging the divide between seemingly unrelated works in the spatial and spectral domains, as well as by linking methods within each domain.
The study will build a unified theoretical framework that encompasses diverse GNNs.
Our research is novel in that it connects disparate GNN models, allowing for direct rethinking and comparison of all GNN models.

\subsection{Graph Neural Networks in Spatial and Spectral Domain}
Over the past several years, GNNs have gained a lot of attention. 
However, the existence of numerous GNNs complicates model selection because they are not easily understood within the same framework.
Specifically, one uses spectral theory to implement early GNNs \cite{hammond2011wavelets,defferrard2016convolutional}, whereas spatial theory is used to propose others \cite{hamilton2017inductive,chen2018fastgcn}. The mismatch inherent to spectral and spatial approaches means that direct comparisons are difficult. Even in each area, there are numerous models, which makes it difficult to examine their strengths and weaknesses.

To untangle the mess, we present a unified framework that connects the spatial and spectral domains and reveals their intricate relationship. Furthermore, both domains' subcategories are proven to have a hierarchical link.
The focus on a unified framework adds to the knowledge of how GNNs operate. 
The goal of this research is to use a combination of spectral graph theory and approximation theory to investigate the relationship between important categories, such as spatial and spectral-based approaches. 
We give a detailed analysis of GNNs' current research findings in this paper, as well as a discussion of trending topics such as over-smoothing. Many well-known GNNs will be used to demonstrate the universality of our architecture. This article's main motivation is twofold:
\textbf{(1) Connecting the spectral and spatial domains.} 
    The fundamental concepts, principles, and physical implications of spectral- and spatial-based GNNs are significantly different due to their distinct features. 
    As a result, we present an overview of the fundamental principles and properties of spectral- and spatial-based GNNs in order to help people better grasp the problems, potential, and necessity of GNNs. 
    Formally, a rigorous equivalence is established, indicating that spatial connection function is comparable to spectral filtering;
\textbf{(2) Dissecting spectral and spatial domains, respectively.}
    In spectral techniques, filtering functions on eigenvalues are examined, and the filtering function choice can be matched with various tactics in approximation theory. While spatial methodologies are used to explore attribute aggregation, which may be understood from the size and direction within a predetermined region.

The structure of the article is summarized as follows: 
Basic concepts, distinctive principles, and properties of graph neural networks are covered in Section \ref{sec:problem}, as well as ways for encoding the graph, spectral-based GNNs, spatial-based GNNs, and essential fundamentals. 
The proposed framework is summarized in Section \ref{sec:overview}, which emphasizes the relevance of hierarchy. 
From Section \ref{sec:spatial} through Section \ref{sec:spectral}, we explore exemplary GNN models in each domain using our proposed taxonomy structure. 
In Section \ref{sec:theory}, we go over the advantages and disadvantages of each domain in detail, as well as practical guidance on GNN model selection. Section \ref{sec:example} also includes a case study of our techniques, demonstrating our proposed framework with current and relevant GNNs concerns. 

\subsection{Related Surveys and Our Contributions}
Existing works can be divided into three groups:
\textbf{Existing Works Group 1 (Comprehensive Collection)}:
Recently, many extensive surveys on graph neural networks have been compiled \cite{zhou2018graph,zhang2018deep,wu2019comprehensive,bronstein2017geometric,hamilton2017representation,zhu2021interpreting}. Instead of studying their hierarchical and underlying mechanisms, most existing surveys focus on gathering newly published works and categorizing them into separate categories. A detailed survey, in particular, provides an overview of many examples of graph neural networks, classifying them as spatial or spectral-based techniques \cite{bronstein2017geometric}. A taxonomy of graph types, training methods, and propagation processes was recently published in \cite{zhou2018graph}. Another survey \cite{zhang2018deep} categorized graph neural network advances as semi-supervised (graph convolution), unsupervised (graph auto-encoder), and latest advancements (graph recurrent neural network and graph reinforcement learning). Graph convolution, graph auto-encoder, graph recurrent neural network, and spatial-temporal graph neural networks are all included in \cite{wu2019comprehensive}. These existing surveys, on the other hand, fail to integrate their categories into a cohesive framework.
\textbf{Existing Works Group 2 (Particular Perspectives)}: 
The second thread of surveys for graph neural networks is from diverse theoretical perspectives. 
For example, in the field of graph neural networks with an attention mechanism, a comprehensive and concentrated survey was undertaken \cite{10.1145/3363574}. Another example demonstrated how many graph neural networks with negative sampling might be merged into an analytical matrix factorization framework \cite{qiu2018network}. One similar study offered a general view proving that network embedding techniques and matrix factorization are equal in terms of two objectives: one for similar nodes and the other for distant nodes \cite{liu2019general}. One specific survey created four benchmark datasets with diverse features and user-friendly interfaces for 10 common algorithms, providing a unified paradigm for systematic categorization and analysis on several existing heterogeneous network embeddings approaches \cite{yang2020heterogeneous}. A recent work analyzed anonymous and degree-aware message-passing to study the distinguishing power of different classes \cite{geerts2021let}.
However, this research is limited to a subset of the GNNs family and lacks a global perspective. 
\textbf{Existing Works Group 3 (Post-Hoc Explanation)}:
Building post-hoc models and then identifying the underlying patterns from a statistical standpoint is another technique to analyze GNNs \cite{leskovec2019gnnexplainer,baldassarre2019explainability,yuan2020explainability,lin2021generative}. Because neural networks are employed without any theories or domain expertise, this methodology is referred to as "black box". For this reason, post-hoc models have the potential to be biased, subject to adversarial attacks, and difficult to verify. Our research focuses on interpretable graph neural networks, which have a strong theoretical foundation.
\begin{table}[h]
\scalebox{0.8}{
\begin{tabular}{|c|c|c|c|c|}
\hline
                    & Existing Works 1 & Existing Works 2 & Existing Works 3 & This Survey \\ \hline
Theoretical Support &    &       \CIRCLE            &   & \CIRCLE \\ \hline
Comprehensiveness   &    \CIRCLE & &    \CIRCLE & \CIRCLE \\ \hline
\end{tabular}}
\caption{Comparing this study with previous studies}
\vspace{-15pt}
\end{table}
Previous surveys either categorize several disparate groups or only address a few GNNs using a certain methodology. Following the overall goals of our framework, we want to create one framework that unifies GNNs across the spatial and spectral domains as well as within each domain via theoretical support.
It should be noted that the majority of the work presented is related to Graph Convolution Networks (GCN) \cite{kipf2016semi}, which is the most common type of GNNs, and that many other varieties of GNNs are still based on GCN. As a result, we do not differentiate between GNNs and GCN in this context and refer to GNNs in the following sections.

\section{Problem Setup and Preliminary}
\label{sec:problem}

In this section, we outline basic concepts, necessary preliminary, and problem setup of learning node-level representation which is the major task in the GNN literature.
A simple graph is defined as $\mathcal{G} = (\mathcal{V}, \mathcal{E})$, where $\mathcal{V}$ is a set of n nodes and $\mathcal{E}$ represents edges.
An entry $v_{i} \in \mathcal{V}$ denotes a node, and $e_{i,j}={\{v_{i}, v_{j}\}} \in \mathcal{E}$ indicates an edge between nodes $i$ and $j$.
The adjacency matrix $\A\in \mathbb{R}^{N\times N}$ is defined by if  
there is a link between node $i$ and $j$, $\A_{i,j} =1$, and else 0. Node features $\X \in \mathbb{R}^{N\times F}$ is a matrix with each entry $x_{i}\in\X$ representing the feature vector on node $i$. Another popular graph matrix is the graph Laplacian which is defined as $\Ll= \D-\A \in \mathbb{R}^{N\times N}$ where $\D$ is the degree matrix. Due to its generalization ability \cite{bollobas2004extremal}
, the symmetric normalized Laplacian is often used, which is defined as $\Ln=\D^{-\frac{1}{2}}\Ll\D^{-\frac{1}{2}}$. Another option is random walk normalization: $\Ln=\D^{-1}\Ll$. 
Note that normalization could also be applied to the adjacency matrix, and their relationship is $\Ln= \I-\An$.
\begin{table}[!hbpt]
    \centering
    \caption{Commonly used notations.}
    \scalebox{0.8}{
    \begin{tabular}{|l|l|}\ChangeRT{2pt}
Notations & Descriptions \\
\ChangeRT{1pt} $\mathcal{G}$ & A graph. \\
\hline$\V$ & The set of nodes in a graph. \\
\hline$\E$ & The set of edges in a graph. \\
\hline$\A, \An$ & The adjacency matrix and its normalization. \\
\hline$\Ll, \Ln$ & The graph Laplacian matrix and its normalization. \\
\hline$v$ & A node $v \in \V .$ \\
\hline$e_{i j}$ & An edge $e_{i j} \in \E.$ \\
\hline$\lambda_{i}\in\LBD$ & Eigenvalue(s). \\
\hline$\UU, \UT$ & Eigenvector matrix and its transpose. \\
\hline$\UU_{i}\in \UU, \ut_{i}\in\UT$ & Single eigenvector and its transpose. \\
\hline $\D$ & The degree matrix of $\mathbf{A}$ and $\mathbf{D}_{i i}=\sum_{j=1}^{n} \mathbf{A}_{i j} .$ \\
\hline $\X \in \mathbf{R}^{N \times d}$ & The feature matrix of a graph. \\
\hline $\Z \in \mathbf{R}^{N \times b}$ & New node feature matrix. \\
\hline $\mathbf{H} \in \mathbf{R}^{N \times b}$ & The node hidden feature matrix. \\
\hline $\mathbf{h}_{v} \in \mathbf{R}^{b}$ & The hidden feature vector of node $v .$ \\
\hline $N$ & node number \\
\hline $b$ & dimension size of hidden feature \\
\hline$\odot$ & Element-wise product. \\
\hline $\mathbf{\Theta},\theta$ & Learnable model parameters. \\
\hline $\Pp(\cdot), \Qq(\cdot)$ & Polynomial function. \\
\hline $\mathcal{N}(v)$ & Directed neighbors of node $v$\\
\ChangeRT{2pt}
\end{tabular}}
\label{tab:notations}
\end{table}
Most GNNs focus on node-level embeddings, learning how a graph signal is modified by a graph topology, and outputting a filtered feature as:
\begin{equation}\small
f: G, \X \rightarrow \Z,
\label{classfication_task}
\end{equation}where we aim to find a mapping which can integrate graph structure and original node features, generating a update node representation $\Z$. ${G}$ represents the graph connectivity, and many options are available as listed in Table \ref{tab:graph_repr}, and most popular are symmetric normalized graph matrices.

\begin{table}[!hbpt]
    \centering
    \caption{Representations for graph topology}
    \scalebox{0.8}{
    \begin{tabular}{|l|l|}\ChangeRT{2pt}
\hline Notations & Descriptions \\
\ChangeRT{1pt} $\A$ & Adjacency matrix \\
\hline$\Ll$ & Graph Laplacian \\
\hline$\tilde{\A}=\A+\I$ & Adjacency with self loop \\
\hline$\D^{-1} \A$ & Random walk row normalized adjacency\\
\hline$\A\D^{-1}$  & Random walk column normalized adjacency\\ 
\hline$\D^{-1/2} \A \D^{-1/2}$ & Symmetric normalized adjacency\\
\hline$\tilde{\D}^{-1} \tilde{\A}$ & Left renormalized adjacency, $\tilde{\D}_{ii}=\sum_{j} \tilde{\A}_{i j}$\\ 
\hline$\tilde{\A}\tilde{\D}^{-1} $ & Right renormalized\\ 
\hline$\tilde{\D}^{-1 / 2} \tilde{\A} \tilde{\D}^{-1 / 2}$ & Symmetric renormalized\\
\hline$(\tilde{\D}^{-1} \tilde{\A})^{k}$ & Powers of left renormalized adjacency\\
\hline$(\tilde{\A}\tilde{\D}^{-1})^{k}$ & Powers of right renormalized adjacency\\
\ChangeRT{2pt}
\end{tabular}}
\label{tab:graph_repr}
\vspace{-12pt}
\end{table}
In this survey, we use the graph Laplacian, adjacency matrix, and their modifications described in Table \ref{tab:graph_repr} to represent a graph. So yet, no experimental or theoretical evidence has been shown that any filter has a consistent advantage \cite{wang2020demystifying}. 
This survey is investigating two specific groups of GNNs, namely spectral- and spatial-based GNNs, which are defined as below:
\begin{definition}[\textit{Spatial Method}]
By integrating graph connectivity ${G}$ and node features $\X$, the updated node representations ($\Z$) are defined as:
\begin{equation}\small
\Z=f({G})\cdot\X,
\end{equation}where ${G}$ is often implemented with $\A$ or $\Ll$ in existing works. Therefore, spatial methods focus on finding a \textbf{node aggregation function} $f(\cdot)$ that learns a proper aggregation with node features $\X$ to obtain a updated node embedding $\Z$.
\label{def:spatial}
\end{definition}


Before introducing another definition, the necessary preliminary background is offered:
\textbf{(1) graph Fourier transform}:
The graph Laplacian $\Ll$ can be diagonalized \cite{shuman2013emerging,zhu2012approximating} as $\Ln = \UU \LBD \UT$, where $\LBD$ is the diagonal matrix whose diagonal elements are the corresponding eigenvalues (i.e., ${\displaystyle \LBD_{ii}=\lambda _{i}}$), and $\UU$ represents eigenvectors.
Further, the graph Fourier transform of a signal $\X$ is defined as $\hat{\X}=\UT \X \in \mathbb{R}^{N\times N}$ and its inverse as $\X=\UU \hat{\X}$. 
\textbf{(2) spectral convolution}: According to the Convolution Theorem (i.e., Fourier transform of a convolution of two signals is the element-wise product of their Fourier transforms) \cite{oppenheim2001discrete}, the convolution is defined in the Fourier domain such that 
\begin{equation*}\small
f_{1}*f_{2}=\UU \left[\left(\UT f_{1} \right) \odot \left(\UT f_{2}\right)\right],
\end{equation*}where $\odot$ is the element-wise product, and $f_{1}/f_{2}$ are two signals defined on the  time or spatial domain.
\begin{definition}[\textit{Spectral Method}]
A node signal $f_{2}=\X$ is filtered by spectral function $\g=\UT f_{1}$ as:
\begin{equation}\small
\g * \X = \UU \left[\g(\LBD)\odot \left(\UT\X\right)\right] = \UU \diag{\g(\LBD)} \UT \X,
\end{equation} where $\g$ is known as \textbf{frequency response function}. If $\g$ is polynomial, rational or exponential function, then we can reduce the equation above to:
\begin{equation}\small
\g * \X = \g(\Ln)\X.
\end{equation}
In short, the objective of spectral methods is to learn a frequency response function $\g(\cdot)$. 
\label{def:spectral}
\end{definition}

The goal of both methods is to learn how to approximate node aggregation or a frequency response function using the data. 
A great deal of approximation techniques are utilized, and thus $f$ or $g$ can be efficiently estimated. 
Approximation theory is a branch of mathematics dedicated to discovering and quantifying the errors caused when functions are approximated using smaller functions.
Despite the fact that polynomials have a more convenient form than rational functions and are more popular due to its efficiency, rational functions are better at approximating functions at singularities and on unbounded domains. The basic characteristics of rational functions are outlined in complex analytic literature \cite{ahlfors1953complex,trefethen2013approximation,pachon2010algorithms,powell1981approximation,cohen2011numerical,petrushev2011rational,achieser2013theory,ziegel1987numerical,boyd2001chebyshev,mason2002chebyshev,remez1934determination}. As an important polynomial approximation, Chebyshev approximation is first introduced as spectral filtering for graph convolution:
A real symmetric graph Laplacian $\Ll$ can be decomposed as $\Ll=\UU \LBD \UU^{-1} =  \UU \LBD \UT$. Chebyshev approximation on spectral filter $\g$ is applied \cite{hammond2011wavelets, defferrard2016convolutional} so that is can be approximated with a polynomials with order k:
\begin{align*}\small
\g * \X =& \UU \g (\LBD) \UT \X &&\\[0.5ex]
\approx&\UU \sum_{k}^{}\theta_{k} T_{k}(\tilde{\LBD}) \UT \X && {(\tilde{\LBD}=\frac{2}{\lambda_{max}}\LBD-\I_{\N})}\\
=&\sum_{k}^{}\theta_{k} T_{k}(\tilde{\Ll}) x &&{(\UU\LBD^{k}\UT=(\UU\LBD\UT)^{k})}
\end{align*}
A most popular graph convolutional network \cite{kipf2016semi} further simplifies this approximation by reducing the order to 1:
\begin{align*}\small
\g * \X \approx& \theta_{0}\I_{\N}x+\theta_{1}\tilde\Ll \X &&({\scriptstyle\text{expand to 1st order)}}\\
=&\theta_{0}\I_{\N}x+\theta_{1}(\frac{2}{\lambda_{max}}\Ll-\I_{\N})) \X &&{\scriptstyle(\tilde\Ll=\frac{2}{\lambda_{max}}\Ll-\I_{\N}))} \\
=&\theta_{0}\I_{\N}x+\theta_{1}(\Ll-\I_{\N})) \X &&{\scriptstyle(\lambda_{max}=2)} \\
=&\theta_{0}\I_{\N}x-\theta_{1} \Dn\A\Dn \X &&{\scriptstyle(\Ll=\I_{\N}-\Dn\A\Dn)} \\
=&\theta_{0}(\I_{\N} + \Dn\A\Dn) \X &&{\scriptstyle(\theta_{0}=-\theta_{1})} \\
=&\theta_{0}(\tilde\D^{-\frac{1}{2}}\tilde\A\tilde\D^{-\frac{1}{2}}) x &&{\scriptstyle(\tilde\A=\A+\I_{\N}, \tilde\D_{ii}=\sum_{j} \A_{ij})}.
\end{align*}
As a result, $\theta_0$ is the only parameter to learn. The learnable parameters in many different GNNs can vary based on the model design.

\section{Framework Overview}\label{sec:overview}

\begin{table}[h]
\caption{Chronological list of notable GNNs in spatial and spectral domains}
\label{tab:milestone}
\scalebox{0.8}{
\begin{tabular}{|p{0.2\textwidth-2\tabcolsep}|p{0.4\textwidth-2\tabcolsep}|p{0.4\textwidth-2\tabcolsep}|}
\ChangeRT{2pt}
            & Spatial & Spectral \\ \hline
Before 2015 
&  ParWalk \cite{wu2012learning}, DeepWalk \cite{perozzi2014deepwalk}, LINE \cite{tang2015line}       
&  Spectral GNN \cite{bruna2014spectral}, ISGNN \cite{henaff2015deep},  Neural graph fingerprints \cite{nfp_neurips15}     \\ \hline
2016        
&  DCNN \cite{atwood2016diffusion}, Molecular Graph Convolutions \cite{gcn_camd16},  PATCHY-SAN \cite{gcn_icml16}     
&    GCN \cite{kipf2016semi}, ChebNet \cite{defferrard2016convolutional}     \\ \hline

2017        
& MPNN \cite{gilmer2017neural}, PGCN \cite{pgcn_neurips17}, GraphSAGE \cite{hamilton2017inductive} 
& MoNet  \cite{monti2017geometric}     
\\ \hline

2018        
& GIN \cite{xu2018how}, Adapative GCN \cite{huang2018adaptive}, Fast GCN \cite{fastgcn_iclr18}  JKNet \cite{xu2018representation}, Large Scale GCN \cite{lgcn_kdd18}
& RationalNet\cite{chen2018rational}, AR \cite{li2018deeper},  CayleyNet \cite{levie2018cayleynets}
\\ \hline

2019        
&  SGCN \cite{wu2019simplifying}, DeepGCN \cite{li2019deepgcns}, MixHop \cite{abu2019mixhop}, PPAP \cite{klicpera2018predict}
&   ARMA \cite{bianchi2019graph},  GDC \cite{klicpera2019diffusion},     EigenPool \cite{eigengcn_kdd19}, GWNN \cite{gwnn_iclr19}, Stable GCNN\cite{sgb_kdd19} 
\\ \hline

2020        
& SIGN \cite{rossi2020sign}, Spline GNN \cite{gssnn_aaai20},  UaGGP \cite{uaggp_aaai20}, GraLSP \cite{gralsp_aaai20}, GraphSAINT \cite{zeng2019graphsaint}, DropEdge \cite{rong2019dropedge}, BGNN\cite{bgnn_ijcai20}, ALaGNN\cite{alagnn_ijcai20} Continuous GNN \cite{cgnn_icml20}, GCNII \cite{chen2020simple}, PPRGo \cite{pprgo_kdd20}, DAGNN \cite{liu2020towards}, H2GCN \cite{h2gcn_neurips20}
&      GraphZoom   \cite{graphzoom_iclr20}
\\ \hline

2021        
&    ADC \cite{adc_neurips21}, UGCN \cite{ugcn_neurips21}, DGC \cite{dgc_neurips21}, E(n)GNN \cite{egnn_icml21}, GRAND \cite{chamberlain2021grand}, C\&S \cite{cs_iclr21}, LGNN \cite{lgnn_ijcai21}
&    Interpretable Spectral Filter \cite{sgf_icml21}, Expressive Spectral Perspective \cite{spexp_iclr21}, S2GC \cite{s2gc_iclr21}, BernNet \cite{he2021bernnet}
\\ \hline

2022        
& GINR\cite{ginr_neurips22}, Adaptive SGC \cite{asgc_neurips22}, PGGNN \cite{pggnn_icml22}, DIMP \cite{dimp_aaai22}
& AGWN \cite{meng2022early}, ChebNetII \cite{chebnetii_neurips22}, JacobiConv \cite{jacobiconv_icml22}, SpecGNN \cite{specgnn_icml22}, G$^2$CN \cite{g2gcn_icml22}, PGNN \cite{pgnn_icml22}, ChebGibbsNet \cite{ChebGibbsNet2023iclr}, SpecFormer\cite{anonymous2023specformer}, SIGN \cite{anonymous2023sign}, Spectral Density \cite{wang2020specdens}
\\ 
\ChangeRT{2pt}
\end{tabular}}
\vspace{-12pt}
\end{table}

The development of GNNs is briefly discussed below with representative studies before we introduce our proposed theoretical framework.
Table \ref{tab:milestone} depicts many sample models that focus on node-level graph convolution. The spectral perspective was previously explored (SGWT \cite{hammond2011wavelets}), and it serves as the technical foundation for all subsequent spectral methods, including spectral convolution and approximation. Researchers continue to add to this thread, demonstrating that spectral methods have the ability to handle graphs (SGNN \cite{bruna2014spectral}, ISGNN \cite{henaff2015deep}, ChebNet \cite{defferrard2016convolutional}). Furthermore, GCN \cite{kipf2016semi}, and GraphSAGE \cite{hamilton2017inductive} create effective training methodologies, gaining considerable attention from various communities. After that, spectral techniques development stagnated, with the exception of a few publications on rational filtering  (RationalNet \cite{chen2018rational}, AR \cite{li2018deeper}, ARMA \cite{bianchi2019graph}). Meanwhile, focus shifts to the spatial domain, which has dominated GNNs to this point. Random walks (ParWalk \cite{wu2012learning}, DeepWalk \cite{perozzi2014deepwalk}, LINE \cite{tang2015line}) and CNN (DCNN \cite{atwood2016diffusion}) were used in early spatial approaches. Following that, MPNN \cite{gilmer2017neural} solidified the message-passing mechanism in spatial techniques. High-order polynomial approximation has been studied \cite{wu2019simplifying,xu2018how,klicpera2019diffusion,rossi2020sign}, but only within the context of ChebNet or DCNN. It's worth noting that while numerous publications described their suggested methods from both spatial and spectral perspectives, only a few GNNs are covered \cite{klicpera2019diffusion,monti2017geometric}. Until recently, spectral research has demonstrated a resurgence.

\begin{figure*}[!t]
    \centering
    \includegraphics[width=0.95\linewidth]{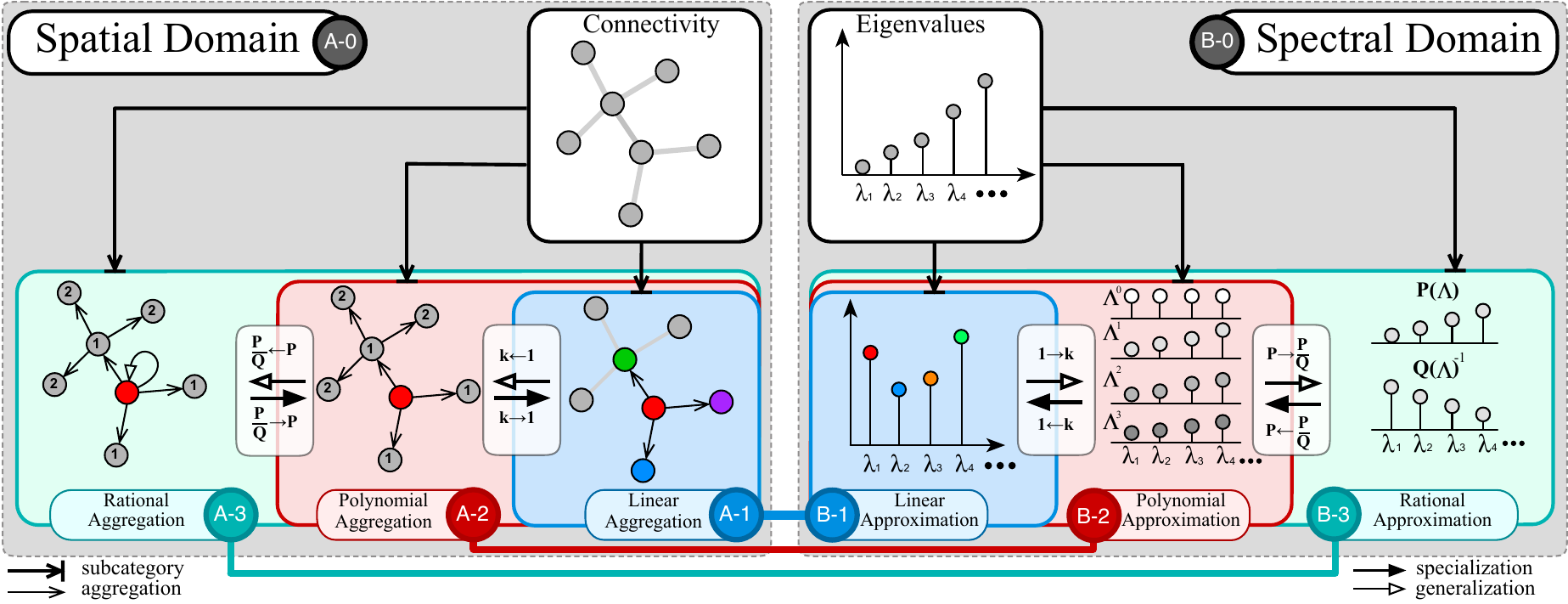}
    \caption{Illustration of major graph neural operations and their relationship. Spatial and spectral methods are divided into three groups, respectively. Groups A-1, A-2, and A-3 are strongly-correlated by generalization and specialization, so are groups B-1, B-3, and B-3. The equivalence among them is marked in the same color.}
    \label{fig:overview}\vspace{-15pt}
\end{figure*}

In this survey, we provide a framework to fully comprehend spectral methods from a spatial perspective and vice versa. A cross-domain perspective is used to integrate spatial and spectral techniques into a coherent framework.
As shown in Figure \ref{fig:overview}, the proposed framework divides GNNs into two domains: spatial (A-0) and spectral (B-0), each of which is further separated into three subcategories (A-1/A-2/A-3 and B-1/B-2/B-3). A-0 is separated into linear (A-1), polynomial (A-2) and rational (A-3) aggregations based on the types of neighbor aggregation (i.e., $f$ in Def. \ref{def:spatial}).  Operations on first-order neighbors only are considered in linear aggregation (A-1), whereas high-order neighbors are incorporated in polynomial aggregation (A-2). In addition, rational aggregation (A-3) includes self-aggregation. 
According to the types of approximation techniques, the spectral methods are divided into linear (B-1), polynomial (B-2), and rational (B-3) approximation depending on the types of frequency filtering (i.e., $\g$ in Def. \ref{def:spectral}). 
In Section \ref{sec:spatial} and \ref{sec:spectral}, each category and subcategory will be explained in detail with examples.


\subsection{Inside the Spatial and Spectral Domain}
The hierarchical link between the spatial and spectral domains is depicted in this subsection. Spatial-based techniques can be divided into three types, with specialization and generalization relationships existing:
\begin{equation*}
\small
    \textsc{(A-1) \underline{Linear Aggregation}} \rightleftarrows \textsc{(A-2) \underline{Polynomial Aggregation}} \rightleftarrows \textsc{(A-3) \underline{Rational Aggregation}},
\end{equation*}where it is a generalization from left to right, and specialization from right to left. 
Specifically, Linear Aggregation (A-1) comprises all algorithms for learning a linear function among neighbors in the first-order. Higher-order neighbors are included in Polynomial Aggregation (A-2), and the order number is defined by the polynomials. Additionally, Rational Aggregation (A-3) utilizes self-aggregation. Since the inclusion of higher-order neighbors causes linear aggregation (A-1) to transform into polynomial aggregation (A-2), and polynomial aggregation (A-2) to transform into rational aggregation (A-3) if self-aggregation is added.
The approaches falling under the general category of spectral analysis can be grouped into three distinct groups:
\begin{equation*}
\small
\textsc{(B-1) \underline{Linear Approximation}} \rightleftarrows \textsc{(B-2) \underline{Polynomial Approximation}} \rightleftarrows \textsc{(B-3) \underline{Rational Approximation}},
\end{equation*} which includes left-to-right generalization and right-to-left specialization. Concretely, (B-1) outlines all models that aggregate frequency components using a linear function, while (B-2) uses polynomial approximation, and (B-3) applies rational approximation. Therefore, (B-1) can be generalized as (B-2) if replacing linear approximation with polynomial approximation, (B-2) is generalized as (B-3) if replacing polynomial approximation with rational approximation. 

\subsection{Between the Spatial and Spectral Domain}\label{subsec:btwn_spa_spec}
In this section, we go through the link between the spatial and spectral domains over the border. There is also equivalency by modifying the analytical form of these subcategories, as shown below. Linear Aggregation (A-1) and Linear Approximation (B-1) are the initial equivalences:
\begin{equation*}\small
    \textsc{(A-1) \underline{Linear Aggregation}} \Leftrightarrow \textsc{(B-1) \underline{Linear Approximation}},
\end{equation*}which means that adjusting weights on neighbors in linear aggregation equates to adjusting weights on frequency components in linear approximation using a linear function. Linear Aggregation (A-1) and Linear Approximation (B-1) have the same linear function and can seamlessly convert to each other in closed form. The main difference between them is that Linear Aggregation (A-1) recovers the signal as a linear function of the frequency component, whereas Linear Approximation (B-1) models the target signal as a linear function of neighbor nodes.
Both Linear Aggregation (A-1) and Linear Approximation (B-1) aggregate the representations of neighbors by tweaking the weights of each neighbor, or uses a linear filter on eigenvalues with a negative slope, i.e., $\g(\LBD)=-\LBD +a$. Because the low-frequency components are given a higher weight by $g$ than their original values, this is referred to as low-pass filtering (i.e., eigenvalues). This group's main advantages are (1) its low computational cost and (2) the large number of real-world scenarios that are subject to the homophily assumption (i.e., neighbors are similar). The fundamental disadvantage is that not every network guarantees homophily.

Polynomial Aggregation (A-2) and Polynomial Approximation (B-2) are identical in terms of actual operation, i.e.,
\begin{equation*}\small
    \textsc{(A-2) \underline{Polynomial Aggregation}} \Leftrightarrow  \textsc{(B-2) \underline{Polynomial Approximation}}.
\end{equation*} This means that in polynomial aggregation, aggregating higher orders of neighbors can be expressed as the sum of different orders of frequency components in polynomial approximation. 
Both use higher-order neighbors in addition to first-order neighbors, increasing the capacity to simulate a more complex relationship among the neighbors. It is theoretically more powerful than (A-1)/(B-1) from a spectral standpoint, because (A-2)/(B-2) is a polynomial approximation as a spectral filter, whereas (A-1)/(B-1) is linear regression. As a result, one flaw is the cost of border neighborhood, which leads to higher computational complexity than (A-1)/ (B-1). Another flaw of them is that if the order is set too large, it will over-smooth (i.e., all nodes will look the same), and there is no golden rule for order size because it is based on data. Note that K-layer (A-1) or (B-1) is equivalent to K-order of (A-2)/(B-2), hence stacking K-layer (A-1) or (B-1) causes over-smoothing (B-1).

\begin{figure*}
    \centering
    \includegraphics[width=.7\linewidth]{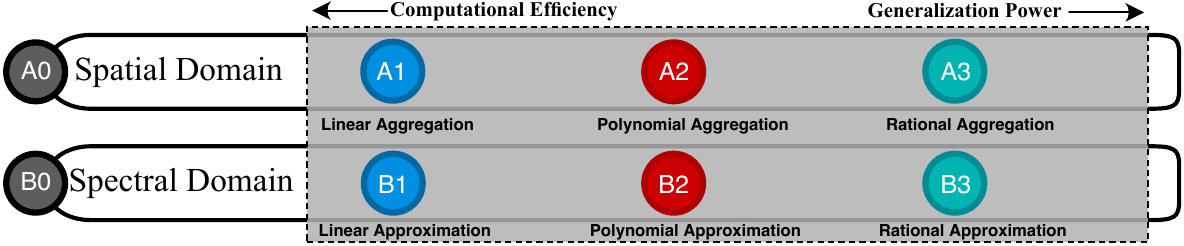}
    \caption{Category and subcategory comparison.}
    \label{fig:cate_comparison}
    \vspace{-15pt}
\end{figure*}

Similarly, the last equivalence relationship is
\begin{equation*}\small
    \textsc{(A-3) \underline{Rational Aggregation}} \Leftrightarrow \textsc{(B-3) \underline{Rational Approximation}},
\end{equation*}in which
rational aggregation defines a label aggregation with self-aggregation, while rational approximation adjusts filter function with rational approximation. Both alleviate the over-smoothing issue by introducing self-aggregation, which limits the intensity of uni-directional aggregation in the spatial domain. From a spectral perspective, this advantage can be interpreted as the superiority of rational approximation (A-3/B-3) over polynomial approximation (A-2/B-2). In particular, the rational approximation is more powerful and precise, particularly when estimating some abrupt signals like discontinuity. \cite{trefethen2013approximation,powell1981approximation,cohen2011numerical,petrushev2011rational,achieser2013theory,boyd2001chebyshev,mason2002chebyshev}
As a result, we may summarize the advantages and disadvantages of each combination as illustrated in Figure \ref{fig:cate_comparison}. The category selection is based on the data complexity and the efficiency required, as there is a trade-off between computational efficiency and generalization capability. 
Table \ref{tab:overall_ab} shows the structure of sections \ref{sec:spatial} and \ref{sec:spectral}, where we will discuss details of these two threads respectively and exemplify using several representative graph neural networks.
The second column denotes spatial perspective (section 4, A-0), which can treat popular GNNs as learning a function of adjacency matrix, or \textit{node aggregation function}. Similarly, the third column means spectral view (section 5, B-0), which sees GNN as learning a function of eigenvalues or frequency response functions.
The cell at the intersection of the second row and second column means that this category of GNNs can be treated as a linear function of adjacency matrix, or say, its node aggregation function is linear. The other intersection cells follow a similar schema.
Note that categories within the same row have the same format and function. For instance, in the second row, A-1 and B-1 share the format of a linear function, but their parameters need not be identical.

\begin{table}[htpb!]
\caption{Structure of section \ref{sec:spatial} and \ref{sec:spectral}.}
\scalebox{0.9}{
\begin{tabular}{|r|c|c|}
\ChangeRT{2pt}
           & \thead{Section \ref{sec:spatial} \colorbox{lightgray}{\textbf{(A-0})} \\ \small{\texttt{Spatial}: \normalfont{function of adjacency matrix}}}  &    
           \thead{Section \ref{sec:spectral} \colorbox{lightgray}{\textbf{({B-0})}} \\\small{\texttt{Spectral}: \normalfont{function of eigenvalues}}}   \\ \hline
Linear\quad $l(\cdot)$     & \thead{sub-section \ref{sec:spatial_a1} \colorbox{lightgray}{\textbf{(A-1})}\\ \small{$l(A)=a_{1}A+a_{0}A^{0}=a_{1}A+a_{0}I$ }}    & \thead{sub-section \ref{sec:spectral_b1} \colorbox{lightgray}{\textbf{(B-1})}\\ \small{$l(\Lambda)=a_{1}\Lambda+a_{0}\Lambda^{0}=a_{1}\Lambda+a_{0}$}}                          \\ \hline
Polynomial\quad $\Pp(\cdot)$ & \thead{sub-section \ref{sec:spatial_a2} \colorbox{lightgray}{\textbf{(A-2})}\\ \small{$\Pp(A)=a_{m}A^{m}+\ldots+a_{k}A^{k}+\ldots+a_{0}A^{0}$}  }    & \thead{sub-section \ref{sec:spectral_b2} \colorbox{lightgray}{\textbf{(B-2})}\\ \small{$\Pp(\Lambda)=a_{m}\Lambda^{m}+\ldots+a_{k}\Lambda^{k}+\ldots+a_{0}\Lambda^{0} $ } }                        \\ \hline
Rational\quad $\frac{\Pp(\cdot)}{\Qq(\cdot)}$   & \thead{sub-section \ref{sec:spatial_a3} \colorbox{lightgray}{\textbf{(A-3})}\\$\frac{\Pp(A)}{\Qq(A)}=\frac{a_{m}A^{m}+\ldots+a_{k}A^{k}+\ldots+a_{0}A^{0}}{a_{m}A^{m}+\ldots+a_{k}A^{k}+\ldots+a_{0}A^{0}}$} & \thead{sub-section \ref{sec:spectral_b3} \colorbox{lightgray}{\textbf{(B-3})}\\$\frac{\Pp(\Lambda)}{\Qq(\Lambda)}=\frac{a_{m}\Lambda^{m}+\ldots+a_{k}\Lambda^{k}+\ldots+a_{0}\Lambda^{0}}{a_{m}\Lambda^{m}+\ldots+a_{k}\Lambda^{k}+\ldots+a_{0}\Lambda^{0}}$} \\ 
\ChangeRT{2pt}
\end{tabular}}
\label{tab:overall_ab}
\vspace{-15pt}
\end{table}

\section{Spatial-based GNNs (A-0)} 
\label{sec:spatial}
In the current literature, spatial approaches such as self-loop, normalization, high-order neighbors, aggregation, and node combination are often explored. 
We established a new taxonomy for spatial-based GNNs based on these operations, dividing them into three separate groupings as below.
    
\subsection{Linear Aggregation (A-1)}\label{sec:spatial_a1}
\begin{figure*}[!t]
    \centering
    \includegraphics[width=0.5\linewidth]{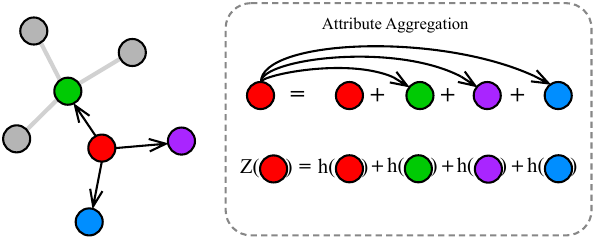}
    \caption{Illustration of A-1: the current node (red) updates itself with its original representation plus neighbors.}
    \label{fig:a1_vis}
    \vspace{-5pt}
\end{figure*}

A lot of research has gone into understanding the aggregation among first order neighbors (i.e., direct neighbors)  \cite{perozzi2014deepwalk,xu2018how,xu2018powerful,gilmer2017neural,hamilton2017inductive,velivckovic2017graph}. The supervisory signal patterns are revealed by altering the weights for the node and its first order neighbors. The revised node embeddings, $\Z(v_i)$, can be represented in the following way:
\begin{equation}\small
    \Z(v_{i}) = \overbrace{\Phi(v_{i}) \h(v_{i}) }^{\text{self node}}+ \overbrace{\sum_{u_{j}\in\mathcal{N}(v_{i})}\Psi(u_{j})\h(u_{j})}^{\text{neighbors' aggregation}}, 
\label{eq:a1}
\end{equation}where $u_{j}$ denotes a neighbor of node $v_{i}$, $\h(\cdot)$ is their representations, and $\Phi/\Psi$ indicate the weight functions. The first item on the right hand side denotes the weighted representation of node $v_{i}$, while the second represents the update from its neighbors. By applying random walk normalization (i.e., dividing neighbors by degree of the current node), Equation \ref{eq:a1} and its symmetric normalization can be written as:
\begin{equation}\small
    \Z(v_{i}) = \Phi(v_{i}) \h(v_{i}) +  
    \sum_{u_{j}\in\mathcal{N}(v_{i})}\Psi(u_{j})\frac{\h(u_{j})}{d_{i}}, \qquad 
    \tilde{\Z}(v_{i}) = \Phi(v_{i}) \h(v_{i}) +  
    \sum_{u_{j}\in\mathcal{N}(v_{i})}\Psi(u_{j})\frac{\h(u_{j})}{\sqrt{d_{i}d_{j}}},
\label{eq:a1_comb}
\end{equation}
where $d_{i}$ represents the degree of node $v_{i}$. Normalization has better generalization capacity, which is not only due to some implicit evidence but also because of a theoretical proof on performance improvement \cite{johnson2007effectiveness}. In a simplified configuration, weights for all the neighbors are the same and is a scalar value $\psi$, while the weight for self node $\phi$ is another scalar value. Therefore, they can be rewritten in matrix form as:
\begin{equation}\small
\Z = \phi \X + \psi\D^{-1}\A\X  = (\phi \I + \psi\D^{-1}\A)\X, \qquad
\tilde{\Z} = \phi \X + \psi\Dsym\A\Dsym\X = (\phi \I + \psi\Dsym\A\Dsym)\X.
\label{eq:a1_mat_comb}
\end{equation} 
Equations \ref{eq:a1_mat_comb} can be generalized as an identical form: 
\begin{equation}\small
\Z =(\phi \I + \psi\tilde{\A})\X,
        \label{eq:a1_final}
\end{equation}
where $\tilde{\A}$ denotes normalized $\A$, which could be implemented by random walk or symmetric normalization. 
As shown in Figure \ref{fig:a1_vis}, the new representation of the current node (in red) is updated as the sum of the previous representations of itself and its neighbors. (A-1) may adjust the weights of the neighbors. 
The following are a few state-of-the-art approaches that were chosen to demonstrate this schema:

\subsubsection{\textbf{Graph Convolutional Network (GCN)}} 
As the one state of the art, GCN \cite{kipf2016semi} adds a self-loop to nodes, and applies a \textit{renormalization} trick which changes degree matrix from $\D_{ii}=\sum_{j}\A_{ij}$ to $\hat{\D}_{ii}=\sum_{j}(\A+\I)_{ij}$. Specifically, GCN can be written as:
\begin{equation}\small
\Z = \hat{\D}^{-\frac{1}{2}}\hat{\A}\hat{\D}^{-\frac{1}{2}}\X = \hat{\D}^{-\frac{1}{2}}(\I+\A)\hat{\D}^{-\frac{1}{2}} \X=(\I+\An)\X,
\label{eq:a1_gcn}
\end{equation}
where $\hat{\A}=\A+\I$, and $\An$ is normalized adjacency matrix with self-loop. Therefore, Equation \ref{eq:a1_gcn} is equivalent to Equation \ref{eq:a1_final} when setting $\phi=0$ and $\psi=1$ with the \textit{renormalization} trick. Besides, GCN takes the sum of each node and average of its neighbors as new node embeddings. Note that the normalization of GCN is different from the others, but the physical meaning is the same.

\subsubsection{\textbf{GraphSAGE}}
Computing intermediate representations of each node and its neighbors, GraphSAGE \cite{hamilton2017inductive} applies an aggregation among its neighbors. Take mean aggregator as example, it averages a node with its neighbors, i.e., 
\begin{equation}\small
\Z(v_{i}) =\operatorname{MEAN}\left(\left\{\h(v_{i})\right\} \cup\left\{\h(u_{j}), \forall u_{j} \in \mathcal{N}(v_{i})\right\}\right), 
\label{eq:a1_sage_intro}
\end{equation}where $\h$ indicates the intermediate representation, and $\mathcal{N}$ denotes the neighbor nodes. Equation \ref{eq:a1_sage_intro} can be written in matrix form after implementing MEAN using symmetric normalization:
\begin{equation}\small
\Z = \Dsym(\I+\A)\Dsym\X=(\I+\An)\X,
\label{eq:a1_sage}
\end{equation} 
which is equivalent to Equation \ref{eq:a1_final} with $\phi=1$ and $\psi=1$. Note that the key difference between GCN and GraphSAGE is the normalization strategy: the former is symmetric normalization with renormalization trick, and the latter is random walk normalization.


\subsubsection{\textbf{Graph Isomorphism Network (GIN)}} Inspired by the Weisfeiler-Lehman (WL) test, GIN \cite{xu2018how} developes conditions to maximize the power of GNNs, proposing a simple architecture, Graph Isomorphism Network (GIN). With strong theoretical support, GIN generalizes the WL test and updates node representations as:
\begin{equation}\small
    \Z=\left(1+\epsilon\right) \cdot \h(v)+\sum_{u_{j}} \in \mathcal{N}(v_{i}) \h(u_{j})=[(1+\epsilon)\I + \A]\X,
\label{eq:a1_gin}
\end{equation} 
which is equivalent to Equation \ref{eq:a1_final} with $\phi=1+\epsilon$ and $\psi=1$. Note that GIN does not perform any normalization.

\subsubsection{\textbf{Graph Attention Model (GAT)}}
GAT \cite{velivckovic2017graph} applies attention mechanism by adjusting neighbors' weights, instead of using uniform weights in many related works:
\begin{equation}\small
      \Z=(W_{att}\otimes\A)\X,
\label{eq:gat_spatial}
\end{equation}
where $W_{att}\in \mathbb{R}^{N\times N}$ is a matrix, $\otimes$ denote element-wise multiplication, and calculated by a forward neural network $W_{att}(i,j) = f(\h_{i}, \h_{j})$ with a pair of node representations as input.  GAT can be treated as learning dynamic weight on the neighbors since their weights are not uniform. MoNet \cite{monti2017geometric} is similar to GAT, since its update follows:
\begin{equation}\small
\Z(v) =\sum_{u \in \mathcal{N}(v)} w_{j}(\mathbf{u}(\h_{i},\h_{j})) \h_{j},
\end{equation}
where $\mathbf{u}$ is a d-dimensional vector of pseudo-coordinates $\mathbf{u}(x, y)$, and 
\begin{equation}\small
w_{j}(\mathbf{u})=\exp \left(-\frac{1}{2}\left(\mathbf{u}-\boldsymbol{\mu}_{j}\right)^{\top} \boldsymbol{\Sigma}_{j}^{-1}\left(\mathbf{u}-\boldsymbol{\mu}_{j}\right)\right),
\end{equation} where $ \mu_{j}$  are learnable  d $\times$ d  and  d $\times$ 1 covariance matrix and mean vector of a Gaussian kernel, respectively. Let $W_{Monet} = w(u(\cdot))$ as a weight function of a pair of node representations representation, then it is also a attention model:
\begin{equation}\small
      \Z=(W_{MoNet}\otimes\A)\X,
\label{eq:monet_spatial}
\end{equation}
These works do not consider updating nodes with their original representations, i.e., $\phi=0$ and $\psi$ is replaced with matrix parameter $W$ in Equation \ref{eq:a1_final}. However, it is easy to extend them with self node.

\subsection{Polynomial Aggregation (A-2)}\label{sec:spatial_a2}
Several research use higher orders of neighbors to integrate deeper structural information \cite{atwood2016diffusion,defferrard2016convolutional,wu2019simplifying,tang2015line,grover2016node2vec}. Because direct neighbors aren't always enough to describe a node's surroundings. Excessive order, on the other hand, generally averages all node representations, resulting in over-smoothing and a loss of emphasis on the immediate neighborhood \cite{li2018deeper}. Many models are motivated by this to fine-tune the aggregation strategy based on different orders of neighbors. As a result, adequate constraint and order flexibility are essential for node representation. Challenging signals, such as Gabor-like filters, have been shown to have a high order of neighbors \cite{abu2019mixhop}.
\begin{figure*}[!t]
    \centering
    \includegraphics[width=0.5\linewidth]{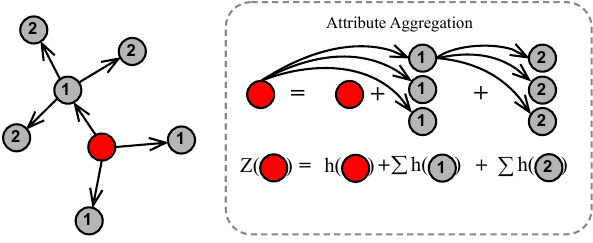}
    \caption{Illustration of A-2: This current node (red) is using its original representation plus its first and second-order neighbors to update itself.}
    \vspace{-10pt}
    \label{fig:a2}
\end{figure*}

Define the shortest distance between node $i$ and $j$ as $d_{G}(i,j)$, and $\partial \mathcal{N}(i, \tau)$ to be the set of nodes $j$ that satisfies $d_{G}(i,j)=\tau$, i.e., $\tau$-order neighbors. Formally, this type of work can be written as:

\begin{equation}\small
    \Z(v_{i}) = \Phi (v_i) \h(v_{i}) 
    +\overbrace{\sum_{u_{j}\in \mathcal{N}(v_i, \tau=1)} \Psi^{(\tau=1)} \h(u_{j})}^\text{1st-order neighbor} +
    \overbrace{\sum_{u_{j}\in \mathcal{N}(v_i, \tau=2)} \Psi^{(\tau=1)} \h(u_{j})}^\text{2nd-order neighbor} 
    +...+
    \overbrace{\sum_{u_{j}\in  \mathcal{N}(v_i, \tau=k)} \Psi^{(\tau=k)} \h(u_{j})}^\text{k-th order neighbor}+...,
\label{eq:a2}  
\end{equation}where $\Psi^{(\tau)}$ indicates a scalar parameter for all $\tau$-order neighbors. Setting the same order neighbors to share the same weights, Equation \ref{eq:a2} can be rewritten in matrix form:
\begin{equation}\small
        \Z= (\phi\I + \sum_{j=1}^{k} \psi_{j}\A^{j}) \X = \Pp_k(\A)\X,
\label{eq:a2_mat}
\end{equation}
where $\Pp_k(\cdot)$ is a polynomial function with order number k. Applying symmetric normalization, Equation \ref{eq:a2_mat} can be rewritten in matrix form as:
\begin{align}
\Z = (\phi \I + \sum_{j=1}^{k}\psi_{i} (\Dsym\A\Dsym)^{j})\X 
    =(\phi \I +\sum_{i=1}^{k} \psi_{i} \An^{i}) \X 
    =(\sum_{i=0} \psi_{i} \An^{i}) \X 
    =\Pp_k(\An)\X,
\label{eq:a2_mat_norm}
\end{align}where $\phi = \psi_{0}$, and $\A$ could also be normalized by random walk normalization: $\An=\D^{-1}\A$. 
As shown in Figure \ref{fig:a2}, the new representation of the current node (in red) is updated as the sum of the previous representations of itself, its first and second-order neighbors. Note that the weights among those representations are learnable.
Several existing works are analyzed below, showing that they are variants of Equation \ref{eq:a2_mat} or \ref{eq:a2_mat_norm}.
 
\subsubsection{\textbf{ChebNet}}
To bridge the gap, spectral convolutional operation is generalized, which requires expensive steps of spectral decomposition and matrix multiplication \cite{bruna2013spectral,henaff2015deep}.
Introducing truncated Chebyshev polynomial for estimating wavelet in graph signal processing, ChebNet \cite{defferrard2016convolutional} embeds a novel neural network layer for the convolution operator. Specifically, ChebNet can be written as:
\begin{equation}\small
    \sum_{k=0}^{K-1} \theta_{k} T_{k}(\tilde{\Ll}) \X= (\tilde{\theta_{0}}\I + \tilde{\theta_{1}}\tilde{\Ll} + \tilde{\theta_{2}}\tilde{\Ll}^{2} +... ) \X,
\end{equation}
where $T_{k}(\cdot)$ denotes the Chebyshev polynomial and $\theta_{k}$ is the Chebyshev coefficient. $\tilde{\theta}$ is the coefficient after expansion and reorganization. Since $\Ln=\I-\An$, we have:
\begin{equation}
   \sum_{k=0}^{K-1} \theta_{k} T_{k}(\tilde{\Ll}) \X=  [\tilde{\theta_{0}}\I + \tilde{\theta_{1}}(\I-\An) + \tilde{\theta_{2}}(\I-\An)^{2} +... ] \X = (\phi \I +\sum_{i=1}^{k} \psi_{i} \An^{i})\X =\Pp_k(\An)\X,
\label{eq:a2_chebnet}
\end{equation}
which is exactly Equation \ref{eq:a2_mat_norm}. The predefined $K$ is the order number of Chebyshev polynomial, and a larger K mean higher approximation accuracy in estimating the function of eigenvalues. Equation \ref{eq:a2_chebnet} shows that K also can be explained as the highest order of the neighbors.  

\subsubsection{\textbf{DeepWalk} }
Applying random walk, DeepWalk \cite{perozzi2014deepwalk} first draws a group of random paths from a graph and applies a skip-gram algorithm to extract node features. Assuming the number of random walk is large enough, the transition probability of random walk on a graph can be represented as:
\begin{equation}\small
   \An=\D^{-1}\A.
\label{eq:a2_deepwalk_transition}
\end{equation} 

Let the window size of skip-gram be $2t+1$ and the current node is the (t+1)-th one along each sampled random walk path, so the farthest neighbor current node can reach is the first one and the last one. 
A node and its neighbors are likely to appear in the same random walk path, and the neighbors follow the transition probability (Equation \ref{eq:a2_deepwalk_transition}) to appear in the same path.
Therefore, the updated representation is as follows:
\begin{equation}\small
    \Z = \frac{1}{t+1}(\I + \An + \An^{2} + ... + \An^{t})\X = \frac{1}{t+1}\Pp_k(\An)\X,
    \label{eq:a2_deepwalk}
\end{equation}where $\I$ means that DeepWalk always considers previous representation of the current node as one element, $\An$ represents the direct neighbors' transition probability, and $\An^{2}$ denotes that of the second-order neighbors. It will have at most $t$-order neighbors depending on the predefined length of the random walk (i.e., $2t+1$).


 
\subsubsection{\textbf{Diffusion convolutional neural networks (DCNN)}} 
DCNN~\cite{atwood2016diffusion} uses a degree-normalized transition matrix (i.e., renormalized adjacency matrix: $\An = \Dn\A$) as graph representation, and performs node embedding update as:
\begin{equation}\small
    \Z=W \odot \An^{*} \X = [w_1, w_2, w_3, ..., w_k]^\intercal\odot [\An, \An^2, \An^3..., \An^k]^\intercal \X,
\label{eq:a2_dcnn}
\end{equation}where $\An^{*}$ denotes a tensor containing the power series of $\An$, and the $\odot$ operator represents element-wise multiplication. It can be transformed as:
\begin{equation}
  \Z=(w_{1}\An + w_{2}\An^{2}+ w_{3}\An^{3} +...w_{k}\An^{k})\X=\Pp_k(\An)\X.
\end{equation}

\subsubsection{\textbf{Scalable Inception Graph Neural Networks (SIGN)}}
By generalizing GCN~\cite{kipf2016semi}, GAT~\cite{velivckovic2017graph} and SGC~\cite{wu2019simplifying}, SIGN~\cite{rossi2020sign} constructs a block linear diffusion operators along with non-linearity. For node-wise classification tasks, SIGN has the form:
\begin{equation}\small
\begin{aligned}
\mathbf{Z} &=\sigma\left(\left[\mathbf{X} \Theta_{0}, \mathbf{A}_{1} \mathbf{X} \Theta_{1}, \ldots, \mathbf{A}_{r} \mathbf{X} \Theta_{r}\right]\right), \\
\mathbf{Y} &=\xi(\mathbf{Z} \Omega),
\end{aligned}
\end{equation}where $[\cdot, \cdot, \ldots,]$ is concatenation, and $r$ denotes the power number. Then, SIGN can be rewritten as:
\begin{equation}\small
    \Z = \left[\mathbf{X} \Theta_{0}, \mathbf{A}_{1} \mathbf{X} \Theta_{1}, \ldots, \mathbf{A}_{r} \mathbf{X} \Theta_{r}\right] \Omega =\omega_{0}\sigma(\mathbf{X} \Theta_{0})+ \omega_{1}\sigma( \mathbf{A}_{1} \mathbf{X} \Theta_{1}), \ldots, \omega_{r}\sigma(\mathbf{A}_{r} \mathbf{X} \Theta_{r}),
\end{equation}where $\sigma(\mathbf{A}_{r} \mathbf{X} \Theta_{r})$ could be treated as refined representation of each order of label aggregation by non-linear function $\sigma$ and fully-connected layer $\Omega$, i.e., $\widehat{\mathbf{A}^{r}\X}$. Replacing $A$ with normalized adjacency matrix, it can be rewritten as:
\begin{equation}\small
    \hat{\Z} =\omega_{0}\widehat{\An^{0}\X}+ \omega_{1}\widehat{\An^{1}\X}, \ldots, \omega_{r}\widehat{\An^{r}\X} = \sum_{r} \omega_{r} \widehat{\An^{r}\X} = \widehat{\Pp(\An)\X}.
\label{eq:a2_sign}
\end{equation}

\subsubsection{\textbf{Graph diffusion convolution (GDC)}}
GDC~\cite{klicpera2019diffusion} defines a generalized graph diffusion via the diffusion matrix:
\begin{equation}\small
    \Z=\sum_{k=0}^{\infty} \theta_{k} \boldsymbol{T}^{k},
\end{equation}where $\theta_{k}$ are the weighting coefficients with 
$\sum_{k=0}^{\infty} \theta_{k}=1, \theta_{k} \in[0,1]$, 
$T$ is a generalized undirected transition matrix which includes the random walk transition matrix $T_{rw}= \A\D^{-1}$, 
and the symmetric transition matrix $T_{sym} = \Dsym\A\Dsym$. In the general case, it can be written as :
\begin{equation}\small
    \Z=\sum_{k=0}^{\infty} \theta_{k} \An^{k} = \Pp(\An).
\label{eq:a2_gdc}
\end{equation}

\subsubsection{\textbf{Node2Vec}} Node2Vec \cite{grover2016node2vec} defines a second-order random walk to control the balance between BFS (breath-first search) and DFS (depth-first search). Consider a random walk that traversed an edge between node $t$ and $v$, denoted as ($t$, $v$), and now it resides at node $v$. Then, the transition probabilities to next stop $x$ from node $t$ is defined as:
\begin{equation}\small
    P(t \rightarrow x)=\left\{
        \begin{array}{ll}{
        \frac{1}{p}} & {\text{if } d(t, x)=0, } \text{ return to the source} \\ 
        {1} & {\text{if } d(t, x)=1, \text{ BSF}} \\ 
        {\frac{1}{q}} & {\text{if } d(t, x)=2, \text{ DFS}},
    \end{array}\right.
\end{equation}
where $d(t, x)$ denotes the shortest path between nodes $t$ and $x$. $d(t, x)$=0 indicates a second-order random walk returns to its source node, (i.e., $t\rightarrow v\rightarrow t$), while $d(t, x)$=1 means that this walk goes to a BFS node, and $d(t, x)$=2 to a DFS node. The parameters $p$ and $q$ control the distribution of the three cases. Assuming the random walk is sufficiently sampled, Node2Vec can be rewritten in matrix form: 
\begin{equation}\small
    \Z = (\frac{1}{p}\cdot\overbrace{\I}^\text{source}+\overbrace{\An}^\text{BFS}+\frac{1}{q}\overbrace{(\An^{2}-\An)}^\text{DFS})\X,
\label{eq:node2vec_mat}
\end{equation} which can be transformed and reorganized as:
\begin{equation}\small
    \Z = [\frac{1}{p}\I+(1-\frac{1}{q})\An+\frac{1}{q}\An^{2}]\X=\Pp_{k=2}(\An)\X,
\label{eq:node2vec_mat_simple}
\end{equation} where transition probabilities $\An=\Drw\A$ is random walk normalized adjacency matrix.

\subsubsection{\textbf{LINE\cite{lin2015learning} / SDNE\cite{wang2016structural}} }

These two models consider first order and second-order neighbors as the constraints for learning node embeddings. first order: the nodes representation is forced to be similar to its neighbors, which is equivalent to:
\begin{equation}\small
    \Z = \An\X.
\end{equation} Second-order: the pair of nodes are forced to be similar if their neighbors are similar, which is equivalent to make second-order neighbors similar, therefore we can get the second-order connectivity by taking the power of the original adjacency: 
\begin{equation}\small
    \Z = \An^2\X.
\end{equation} Then the final learned node embeddings are formulated as:
\begin{equation}\small
    \Z = \An\X+\alpha\An^2\X=\Pp_{k=2}(\An)\X.
    \label{eq:a2_line_sdne}
\end{equation}

Since LINE uses concatenation between the representations constrained by first and second-order, $\alpha=1$; For SDNE, $\alpha$ is pre-defined.

\subsubsection{\textbf{Simple Graph Convolution (SGC)}} To reduce the computational overhead, SGC \cite{wu2019simplifying} removes non-linear function between neighboring graph convolution layers, and combines graph aggregation in one single layer:
\begin{equation}\small
    \Z = \An^{k}\X
\end{equation} where $\An$ is renormalized adjacency matrix, i.e., $\An= \Dnsym\A\Dnsym$, where $\Dnsym$ is degree matrix with self loop. Therefore, it can be easily rewritten as:
\begin{equation}\small
    \Z = (0\cdot\I + 0\cdot \An + 0\cdot\An^{2} +...+1\cdot\An^{k})\X = \Pp_k(\An) \X,
    \label{eq:a2_sgc}
\end{equation}which only has the highest order term.



\subsection{Rational Aggregation (A-3)}\label{sec:spatial_a3}
Most works merely consider label propagation from the node to its neighbors (i.e., gathering information from its neighbors) but ignore self-aggregation. Self-aggregation means that labels or attributes can be propagated back to themselves or restart propagating with a certain probability. This reverse behavior can avoid over-smoothing issue \cite{klicpera2018predict}. Note that Polynomial Aggregation (A-2) may manually change the order number to relieve the over-smoothing issue, but Rational Aggregation (A-3) can do so automatically. Theoretically, rational function approximation is more effective than polynomial and has been researched in machine learning problems \cite{telgarsky2017neural,petrushev2011rational,boulle2020rational}.
Several works use a rational function on the adjacency matrix to perform self-aggregation, either explicitly or implicitly \cite{chen2018rational,trimmel2022era,klicpera2018predict,Li_2019_CVPR,7131465,7581108,levie2018cayleynets,bianchi2019graph}. 

\begin{figure*}[!t]
    \centering
    \includegraphics[width=0.5\linewidth]{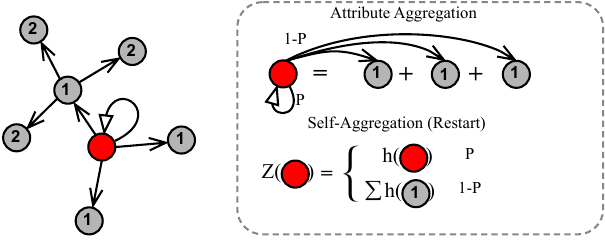}
    \caption{Illustration of A-3: These current nodes (red) are using the representation that predates this iteration and the surrounding nodes to compute the total. In A-3, the ratio of the original representation remains stable, whereas A-1 dose not control the ratio.}
    \vspace{-10pt}
    \label{fig:a3}
\end{figure*}

Because generic label propagation is achieved by multiplying the graph Laplacian, self-aggregation may be achieved by multiplying the inverse graph Laplacian as follows:
\begin{equation}\small
    \Z =  \Pp_m(\An)\Qq_n(\An)^{-1}\X  =\frac{\Pp_m(\An)}{\Qq_n(\An)}\X,
\label{eq:a3_mat}
\end{equation}where $\Pp$ and $\Qq$ are two different polynomial functions, and the bias of $\Qq$ is often set to 1 to normalize the coefficients.
As shown in Figure \ref{fig:a3}, the new representations of the current node (in red) are updated as the previous one with probability P, and as that of neighbors with probability (1-P). The difference of A-3 beyond A-2 is that A-3 can avoid over-smoothing issue in an automatic manner \cite{li2018deeper,zhao2019pairnorm}. Over-smoothing issue happens when GNNs go deep, which would drive node features to a stationary point and average all the information from raw node representations. 
Graph convolution can be described as an optimization problem \cite{Li_2019_CVPR,nt2019revisiting,zhao2019pairnorm, zhu2021interpreting}, e.g., (1) minimizing the supervised loss and (2) keeping the local neighborhood similar:
\begin{equation}\small
    \Z=\underset{\Z}{\arg \min }\{\underbrace{\|\Z-Y\|_{2}^{2}}_{\text {(1) supervised loss}}+\underbrace{\alpha \operatorname{Tr}\left(\Z^{\top} \Ll \Z\right)}_{\text {(2) neighborhood regularization}}\},
\end{equation}where $\alpha$ is the controlling weight between the two constraints. The problem has analytical solution: 
\begin{equation}\small
    \Z=(\I+\alpha \Ln)^{-1} Y=((1+\alpha)\I-\An)Y.
\end{equation}

However, because $alpha$ increases with the number of times graph convolution is done, it is prone to over-smoothing. Over-smoothing is addressed in a variety of ways \cite{huang2020combining,rossi2020sign,rong2019dropedge,xu2018representation,chen2020simple}, including Rational Aggregation (A-3), which does so by retaining a portion of the original representation no matter how many iterations it does, greatly reducing over-smoothing.

\subsubsection{\textbf{Auto-Regressive}}
Label propagation (LP) \cite{zhu2003semi,zhou2004learning,bengio200611} is a widely used methodology for graph-based learning. The objective of LP is two-fold: one is to extract embeddings that match with the node label, the other is to become similar to neighboring vertices. The label can be treated as part of node attributes, so we have:
\begin{equation}\small
    \Z =( \I + \alpha \Ln )^{-1} \X = \frac{\I }{\I +\alpha (\I -\An )} \X = \frac{\I }{(1+ \alpha )\I - \alpha \An} \X,
\end{equation}
which is the closed-form solution and also equivalent to the form of Equation \ref{eq:a3_mat}, i.e., $\Pp=\I$ and $\Qq=(1+\alpha)\I- \alpha \An$.

\subsubsection{\textbf{Personalized PageRank (PPNP)}} Obtaining node's representation via teleport (restart), PPNP \cite{klicpera2018predict,bojchevski2020scaling,ying2018graph} keeps the original representation (self-aggregation) $\X$ with probability $\alpha$. Therefore, 1-$\alpha$ is the probability of performing the normal label propagation:
\begin{equation}\small
    \Z=\alpha\left(\I-(1-\alpha) \An\right)^{-1} \X = \frac{\alpha}{\I-(1-\alpha) \An} \X,
\label{eq:ppnp}
\end{equation}where $\An=\Drw\A$ is random walk normalized adjacency matrix with self-loop. Equation \ref{eq:ppnp} is with a rational function whose numerator is a constant.

\subsubsection{\textbf{ARMA filter}} ARMA \cite{bianchi2019graph} filter approximates any desired filter response function with updates as:
\begin{equation}\small
    \Z=\frac{b}{\I-a\An}\X.
\label{eq:a3_arma}
\end{equation} Note that ARMA filter is an unnormalized version of PPNP. When a+b=1, ARMA becomes PPNP.

\subsubsection{\textbf{ParWalks}}
A partially absorbing random walk is a second-order Markov chain with partial absorption at each state. \cite{wu2012learning} shows that with proper absorption, the absorption probabilities can well capture the global graph structure. Note that the concept "absorption" in \cite{wu2012learning} is similar to "teleport" or "restart" in PPNP \cite{klicpera2018predict}. ParWalks defines the aggregation as:
\begin{equation}\small
    p_{i j}=\left\{\begin{array}{ll}
\frac{\alpha_{i}}{\alpha_{i}+d_{i}}, & i=j \\\\
\frac{w_{i j}}{\alpha_{i}+d_{i}}, & i \neq j,
\end{array}\right.
\label{eq:parwalk_prop}
\end{equation}where $\alpha$ is defined as a variable to control the level of absorption, $w_{ij}$ and $d_{i}$ indicate non-negative matrix of pairwise affinities between vertex $i$ and $j$, and degree of vertex $i$, respectively. 

\begin{equation}\small
    \Z=\frac{\alpha\I}{\alpha + \Ln}\X=\frac{\alpha}{\alpha\I + \I-\An}\X,
\end{equation}
where $\alpha$ is redefined as a regularizer in the original paper \cite{wu2012learning}. 
When $\alpha=1$, all nodes follow the same absorbing behavior. Otherwise, each node has an independent absorbing policy. 
Also, ParWalks model is equivelent to ARMA filter ($a=b=\frac{1}{2}$) when $\alpha=1$ and with normalized Laplacian:
\begin{equation}\small
    \Z=\frac{\I}{\I + \Ln}\X = \frac{\I}{\I +(\I-\An)}\X=\frac{\frac{1}{2}\I}{\I -\frac{1}{2}\An}\X.
\end{equation}

The author also discussed the over-smoothing issue: when
$\LBD=\I$ and as $\alpha\rightarrow 0$, a ParWalk would converge to the constant distribution $1/n$, regardless of the starting vertex.

\subsubsection{\textbf{RationalNet}} To leverage higher order of neighbors, RationalNet \cite{chen2018rational} proposes a general rational function with a predefined order number, and it is optimized by Remez algorithm. The analytic form is exactly Equation \ref{eq:a3_mat}. The major difference beyond PPNP or ARMA filter is that RationalNet generalized them, and the order can be any number.

\begin{figure}[!t]
    \centering
    \includegraphics[width=0.35\linewidth]{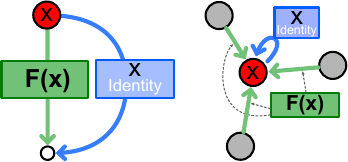}
    \caption{Left: Residual Learning $x'=F(x)+x$; Right: Rational Aggregation: $x'=F(x)+x$}
    \label{fig:a3_rat_as_res}
    \vspace{-10pt}
\end{figure}

\vspace{5pt}
\noindent\textbf{Remark:} 
The optimization towards rational aggregation (A-3) is exactly the same as the \textit{residual learning} that was first and widely used in image recognition \cite{he2016deep}. As shown in the work PPAP \cite{perozzi2014deepwalk}, the author proposed an iterative algorithm called APPAP, which is 
\begin{equation}
   \Z^{(k+1)}=(1-\alpha) {\tilde{A}} \Z^{(k)}+ \alpha \Z^{0}=\overbrace{\sum_{i}^{k}\alpha(1-\alpha)^{i} {\tilde{A}}^{i} \Z^{(0)}}^{F(x) \text{ where } x=\Z^{(0)}}+ \overbrace{\alpha \Z^{0}}^{\text{identity $x$}}
   \label{eq:a3_rat_as_res}
\end{equation}where $\Z^{k}$ means the intermediate representations at $k$-th layer. This format is exactly the same as residual learning as illustrated in Figure \ref{fig:a3_rat_as_res}: sum of multiple graph convolutions serve as $F(x)$, and each time identify input will be added. The slight difference is that Equation \ref{eq:a3_rat_as_res} has normalized the weights, i.e., $(1-\alpha)+\alpha=1$.
Because Rational Aggregation (A-3) includes the inverse of a matrix, it has a high computational cost. Iterative methods are commonly used to efficiently determine the inverse of a matrix \cite{bianchi2019graph,klicpera2018predict,levie2018cayleynets}.
In this subsection, we only demonstrate one layer or iteration of Auto-Regression, PPNP, ARMA, and ParWalks. Multiple iterations will result in a more sophisticated rational function.
Higher orders are achieved by several iterations or layers.
The main distinction between rational and polynomial aggregation is whether or not the inverse graph Laplacian polynomial exists. In each cycle of rational aggregation, a fixed ratio for the original representation is always reserved, whereas polynomial aggregation does not.
On the other hand, calculating the inverse of the graph Laplacian is expensive, making iterative fashion a key object in online learning algorithms \cite{herbster2005online}. By leveraging the concept of conductance, with $f$ as a heat distribution over the vertexes, $\Ll(f)$ indicates the flux induced by $f$ over the graph. Then based on the representer theorem \cite{argyriou2009there,scholkopf2001generalized}, $f(\mathcal{V}_{i})=\Ll^{-1}(\Ll(f))$ could be interpreted as the heat at each vertex been expressed concerning or derived from the flux through every vertex. Thus, when $\Ll$ sends a heat distribution f over each node to flux through each vertex, $\Ll^{-1}$ sends some of the fluxes over the graph back to the original heat distribution (i.e., keep part of fluxes itself). Going back to the graph learning application, we first translate our updated ``heat distribution'' to flux through all of those nodes by calculating $\Pp(\Ll(f))$. M-th degree of $\Pp(\cdot)$ means that each vertex can update M-th neighbors at most. Then using another updated flux in the reverse direction, $\Qq(\Ll(f))^{-1}$ will adjust or reduce flux within N-th neighbors.
Polynomial aggregation with more layers or a higher degree tends to involve more neighbors, increasing capacity. When utilizing too many layers or degrees, over-smoothing is almost always unavoidable (e.g., all nodes are similar). However, unless all of the layers or degrees are tried, determining the appropriate number of layers or degrees is difficult. 
The over-smoothing problem is largely overcome by the ``sending back'' (i.e., teleport) behavior of rational aggregation, in which the out-degree flux is restrained even if excesses of graph convolutional layers or approximation degrees are added.\cite{klicpera2018predict}.




Three groups of spatial methods introduced above (i.e., A-1, A-2, A-3) are strongly connected under \textit{generalization} and \textit{specialization} relationship, as shown in Figure \ref{fig:overview}. \textbf{Generalization:} By adding more neighbors of higher rank, Linear Aggregation (A-1) can be expanded to Polynomial Aggregation (A-2). By adding reverse aggregation, Polynomial Aggregation (A-2) can be advanced to Rational Aggregation (A-3);  \textbf{Specialization:} Linear Aggregation (A-1) is a special case of Polynomial Aggregation where the order is set to 1. (A-2). Rational Aggregation (A-3) degenerates into Polynomial Aggregation when reverse aggregation is removed (A-2).




\begin{table}[]
\caption{Summary of Representative GNNs}
\scalebox{0.8}{
\begin{tabular}{|l|l|l|}\ChangeRT{2pt}
             & (\textbf{A-1}) linear function of $\A$            & (\textbf{B-1}) linear function of $\mathbf{\Lambda}$  \\ \hline
\textbf{GCN}          & $\I+\An$                                        & $2-\mathbf{\Lambda}$           \\ \hline
\textbf{GraphSAGE}    & $\hat{\D}^{-1}+\An$                             & $2-\mathbf{\Lambda}$           \\ \hline
\textbf{GIN}         & $(1+\epsilon)\I + \A$                           & $2+\epsilon-\mathbf{\Lambda}$  \\ \ChangeRT{1pt}
             & (\textbf{A-2}) polynomial function of $\A$           & (\textbf{B-2}) polynomial function of $\mathbf{\Lambda}$  \\ \hline
\textbf{ChebNet}      & $\phi \mathbf{I}+\sum_{i=1}^{k} \psi_{i} \tilde{\mathbf{A}}^{i} $                                                                                                  & $\tilde{\theta}_{0} \cdot 1+\tilde{\theta}_{1} \mathbf{\Lambda}+\tilde{\theta}_{2} \mathbf{\Lambda}^{2}+\ldots$             \\ \hline
\textbf{DeepWalk}     & $\frac{1}{t+1}\left(\mathbf{I}+\tilde{\mathbf{A}}+\tilde{\mathbf{A}}^{2}+\ldots+\tilde{\mathbf{A}}^{t}\right)$                   &  $\frac{1}{t+1}
                [\ldots+\left((-1)^{t-1}+\left(\begin{array}{c}1 \\ t\end{array}\right)(-1)^{t-1}\right)+\ldots]$ 
                \\ \hline
\textbf{DCNN}         & $\psi_{1} \tilde{\mathbf{A}}+\psi_{2} \tilde{\mathbf{A}}^{2}+\psi_{3} \tilde{\mathbf{A}}^{3}+\ldots$                                                                                    &   $\theta_{1} \mathbf{\Lambda}+\theta_{2} \mathbf{\Lambda}^{2}+\theta_{3} \mathbf{\Lambda}^{3}+\ldots$             \\ \hline
\textbf{GDC}         & $\psi_{1} \tilde{\mathbf{A}}+\psi_{2} \tilde{\mathbf{A}}^{2}+\psi_{3} \tilde{\mathbf{A}}^{3}+\ldots$                                                                                    &   $\theta_{1} \mathbf{\Lambda}+\theta_{2} \mathbf{\Lambda}^{2}+\theta_{3} \mathbf{\Lambda}^{3}+\ldots$             \\ \hline
\textbf{Node2Vec}     & $\frac{1}{p} \mathbf{I}+\left(1-\frac{1}{q}\right) \tilde{\mathbf{A}}+\frac{1}{q} \tilde{\mathbf{A}}^{2}$                                                                                    &    $\left(1+\frac{1}{p}\right)-\left(1+\frac{1}{q}\right) \mathbf{\Lambda}+\frac{1}{q} \mathbf{\Lambda}^{2}$            \\ \hline
\textbf{LINE/SDNE}         & $\psi_{1} \tilde{\mathbf{A}}+\psi_{2} \tilde{\mathbf{A}}^{2}$                                                                                    &   $\theta_{1} \mathbf{\Lambda}+\theta_{2} \mathbf{\Lambda}^{2}$             \\ \hline
\textbf{SGC}          & $0 \cdot \mathbf{I}+0 \cdot \tilde{\mathbf{A}}+0 \cdot \tilde{\mathbf{A}}^{2}+\ldots+1 \cdot \tilde{\mathbf{A}}^{K}$                                                                                    &      $\left(\begin{array}{c}K \\ 0\end{array}\right)+\left(\begin{array}{c}K \\ 1\end{array}\right) \mathbf{\Lambda}^{1}+\left(\begin{array}{c}K \\ 2\end{array}\right) \mathbf{\Lambda}^{2}+\cdots+\mathbf{\Lambda}^{n}$          \\ \ChangeRT{1pt}
             & (\textbf{A-3}) rational function of $\A$            & (\textbf{B-3}) rational function of $\mathbf{\Lambda}$  \\ \hline
\textbf{Auto-Regress} & $\frac{\I}{(1+\alpha) \I-\alpha \An}$                                                                                    &   $\frac{1}{1+\alpha(1-\mathbf{\Lambda})}$             \\ \hline
\textbf{PPNP }        & $\frac{\alpha}{\I-(1-\alpha) \An}$                                                                                    &     $\frac{\alpha}{\alpha \mathbf{I}+(1-\alpha) \mathbf{\Lambda}}$           \\ \hline
\textbf{ARMA}         & $\frac{b}{(\I-a \An)}$      &   $\frac{b}{(1-a+a \mathbf{\Lambda})}$             \\ \hline
\textbf{ParWalk}         & $\frac{\alpha}{\alpha\I + \I-\An}$      &   $\frac{\beta}{\beta + \LBD}$       \\  
             \ChangeRT{2pt}
\end{tabular}}
\label{tab:eq_summary}
\end{table}

\section{Spectral-based GNNs (B-0)} 
\label{sec:spectral}
The use of eigen-decomposition and analysis of the weight-adjusting function (i.e., frequency filter function or frequency response function) on eigenvalues of graph matrices are both parts of graph spectral theory. In spectral-based GNNs (B-0), weights are applied to frequency components (eigenvectors) in order to recover the target signal using the filter's output. Accordingly, we propose a new taxonomy for graph neural networks, dividing spectral-based GNNs into three subgroups depending on the types of response filtering functions.
In addition, the same set of representative models discussed in Section \ref{sec:spatial} will be analyzed under spectral view. To facilitate comprehension of the analysis, their spatial and spectral analytical forms are listed in Table \ref{tab:eq_summary}. The detailed transformation of equations in category B-0 is deferred to the appendix.

\subsection{Linear Approximation (B-1)}\label{sec:spectral_b1}
\begin{figure*}[!t]
    \centering
    \includegraphics[width=0.55\linewidth]{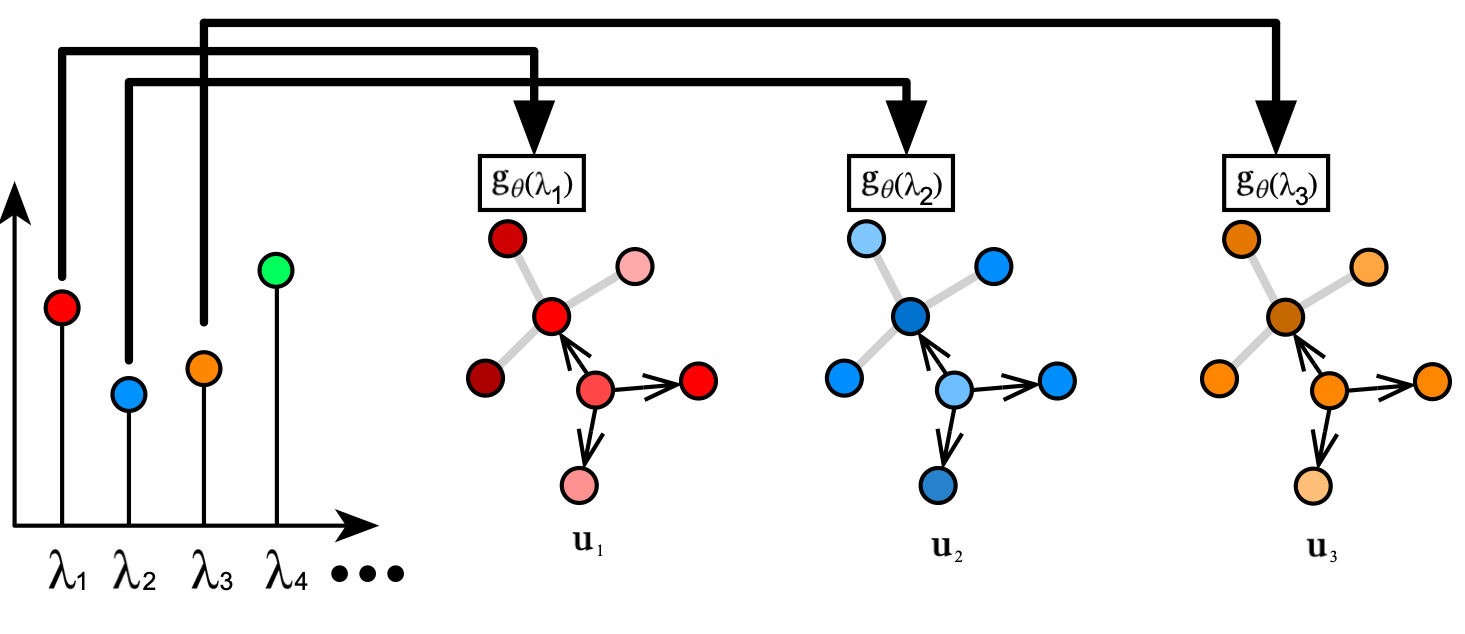}
    \caption{Illustration of B-1: A linear function $g$ maps the eigenvalues to new values.}
    \label{fig:b1}
    \vspace{-8pt}
\end{figure*}

Changing the weights of frequency components in the spectrum domain has been the subject of several research. 
The filter function's objective is to suit the intended output by adjusting eigenvalues. Many of them have been shown to be low-pass filters \cite{Li_2019_CVPR}, which implies that only low-frequency components are highlighted, i.e., the first few eigenvalues are increased, while the rest are decreased. 
There are several studies that may be understood as changing frequency component weights during aggregation. A linear $\g$ function is used in particular: 
\begin{equation}\small
    \Z = (\sum_{i=0}^{l} \theta_{i}\lambda_{i} \uu_{i}\ut_{i} )\X= \UU \g_{\theta}(\LBD) \UT\X,
\end{equation} where $\uu_{i}$ is the i-th eigenvector, and $\g$ is \textit{frequency filter function} or \textit{frequency response function} controlled by parameters $\theta$, with selected $l$ lowest frequency components. The goal of $\g$ is to change the weights of eigenvalues to fit the target output. 
As shown in Figure \ref{fig:b1}, B-1 updates the weights of eigenvectors ($\uu_{1}, \uu_{2}, \uu_{2} \ldots$) as $\g_{\theta}(\lambda)$ which is a linear function. 
Several state-of-the-art methods introduced in Section \ref{sec:spatial} are analyzed to provide a better understanding of this scheme.

\vspace{5pt}
\noindent\textbf{Remark:}
The aforementioned methods apply linear low-pass filtering, and the only difference among them is that the bias is different (i.e., 2 for GCN, 2 for GraphSAGE, and 2+$\epsilon$ for GIN). Therefore, we study the influence of bias on the filter function, and define a metric:
\begin{equation}\small
    w(\lambda_{i}) = \frac{\left|bias-\lambda_{i}\right|}{\sum_{j} \left|bias-\lambda_{j}\right|},
\end{equation}which indicates the overall proportion change of each eigenvalue after applying the response function. A large adjusted value means that the filtering will enlarge the influence of the corresponding eigenvector. The range of the eigenvalue is in [0, 2) for the normalized Laplacian matrix \cite{chung1997spectral}. If let $bias$ be larger than or equal to 2, we have:
\begin{equation}\small
    w(\lambda_{i})
    =\frac{bias-\lambda_{i}}{N\cdot bias-\sum_{j}\lambda_{j}}
    =\overbrace{\frac{-1}{N\cdot bias-\sum_{j}\lambda_{j}}}^{\text{slope}}\lambda_{i}+\overbrace{\frac{bias}{N\cdot bias-\sum_{j}\lambda_{j}}}^{\text{intercept}},
\end{equation}
when  $bias$ is larger or equal than 2, the slope is negative, which means that the filter function is low-pass filtering: as the bias increases, the slope becomes larger, and larger weights are assigned to low-frequency spectral components. Therefore, the bias of all studies in this subsection is larger or equal to 2.


\subsection{Order of Approximation (B-2)}\label{sec:spectral_b2}
Considering higher order of frequency, filter function can approximate any smooth filter function, because it is equivalent to applying the polynomial approximation. Therefore, introducing higher-order of frequencies boosts the representation power of filter function in simulating spectral signal.
Formally, this type of work can be written as:
\begin{equation}\small
    \Z = (\sum_{i=0}^{l} \sum_{j=0}^{k}\theta_{j}\lambda_{i}^{j} \uu_{i}\ut_{i})\X= \UU \Pp_{\theta}(\LBD) \UT\X,
\label{eq:spectral_b2_mat}
\end{equation} where $\g(\cdot)=\Pp_{\theta}(\cdot)$ is a polynomial function.
\begin{figure*}[!t]
    \centering
    \includegraphics[width=0.55\linewidth]{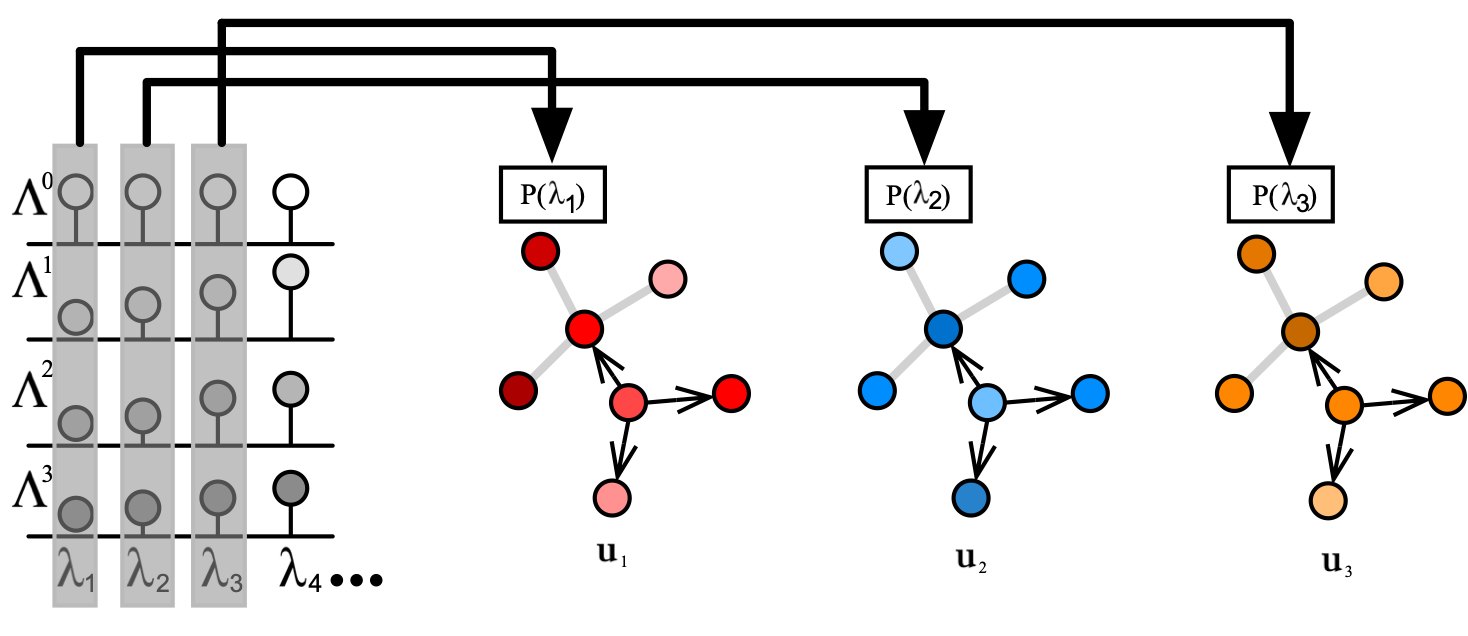}
    \caption{Illustration of B-2: A polynomial function $\Pp$ maps the eigenvalues to new values.}
    \label{fig:b2}
    \vspace{-10pt}
\end{figure*}
As shown in Figure \ref{fig:b2}, B-2 updates the weights of eigenvectors ($\uu_{1}, \uu_{2}, \ldots$) as $\Pp_{\theta}(\lambda)$ which is a polynomial function.

\begin{figure}[h]
	\centering
    \includegraphics[width=0.7\linewidth]{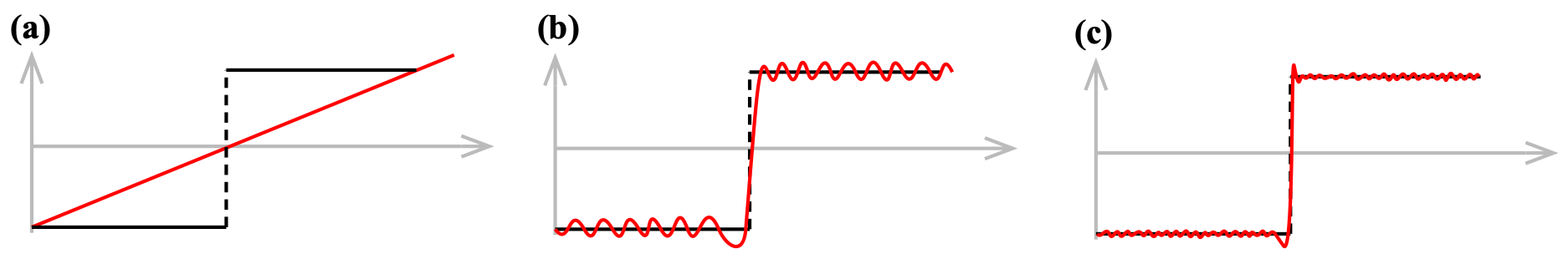}
	\caption{Approximation for $sign(x)$ (in black): (a) linear approximation (b) polynomial approximation with low orders, (c) polynomial approximation with high orders.}
	\label{fig:b2_analysis}
    \vspace{-10pt}
\end{figure}

\vspace{5pt}
\noindent\textbf{Remark:}
Polynomial approximation, in theory, gets more accurate as the order grows \cite{ahlfors1953complex,trefethen2013approximation,pachon2010algorithms,powell1981approximation,cohen2011numerical}. It's worth noting that Linear Approximation (B-1) can be thought of as a polynomial approximation of order 1. We look into polynomial approximation on the $sign(x)$ function, comparing and contrasting all of the cases in Polynomial Approximation (B-2). Because it is difficult for any straight line to suit a jump signal, as shown in Figure \ref{fig:b2_analysis}a, linear functions cannot accurately approximate $sign(x)$. The situation improves dramatically when polynomial approximation is used, as demonstrated in Figure \ref{fig:b2_analysis}b. The variance will be greatly decreased if the order of the polynomial function is increased (Figure \ref{fig:b2_analysis}c). 
To recapitulate, higher-order polynomial approximation is more accurate than lower-order polynomial approximation, but it comes at the expense of increased computational complexity. 
Node2Vec/LINE/SDNE with an order of 2 have lesser approximation power than those with more than 2 layers/orders because the latter's order is predefined and can be as large as possible (e.g., ChebNet  \cite{defferrard2016convolutional}, DeepWalk \cite{perozzi2014deepwalk}, Diffusion CNN \cite{atwood2016diffusion}, Simple Graph Convolution \cite{wu2019simplifying}).

\subsection{Approximation Type (B-3)}\label{sec:spectral_b3}
\begin{figure*}[!t]
    \centering
    \includegraphics[width=0.55\linewidth]{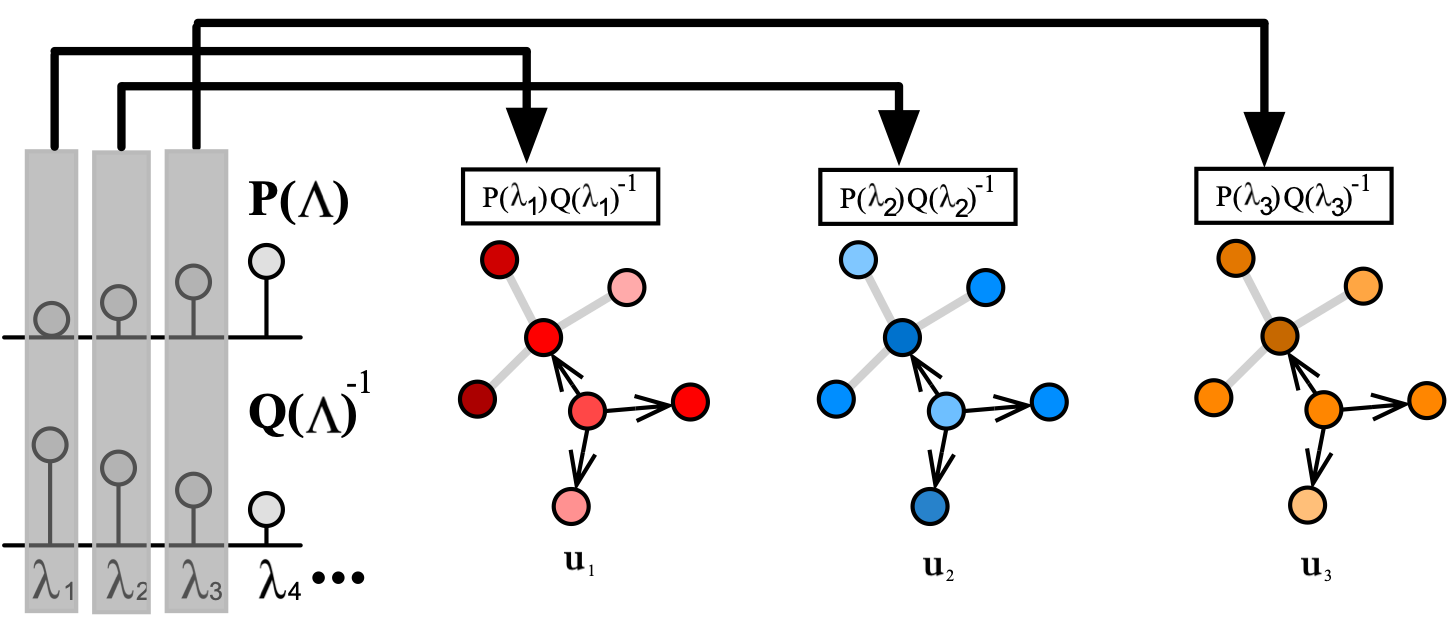}
    \caption{Illustration of B-3: A rational function maps the eigenvalues to new values.}
    \label{fig:b3}
\end{figure*}

Despite its widespread use and experimental success, polynomial approximation only works when applied to a smooth spectral signal. Real-world signals, on the other hand, cannot be guaranteed to be smooth. As a result, the rational approximation is employed to improve the accuracy of non-smooth signal modeling. An example of a rational kernel-based technique is as follows:
\begin{equation}\small
    \Z =(
        \sum_{i}^{l} 
        \frac{\mathlarger\sum_{j=0}^{k}\theta_{j}\lambda_{i}^{j}}
             {\mathlarger\sum_{m=1}^{n}\phi_{m}\lambda_{i}^{m}+1}   
        \uu_{i}\ut_{i}
    )\X
    = \UU \frac{\Pp_{\theta}(\LBD)}{\Qq_{\phi}(\LBD)} \UT\X,
    \label{eq:b3}
\end{equation}where $\g(\cdot)=\frac{\Pp_{\theta}(\cdot)}{\Qq_{\phi}(\cdot)}$ is a rational function, and $\Pp, \Qq$ are independent polynomial functions. Spectral methods process graph as a signal in the frequency domain. 
As shown in Figure \ref{fig:b3}, B-3 updates the weights of eigenvectors ($\uu_{1}, \uu_{2}, \ldots$) as $\g_{\theta}(\lambda)$ which is a rational function. 

\begin{figure}[h]
    \begin{subfigure}
        \centering
        \includegraphics[width=0.2\textwidth]{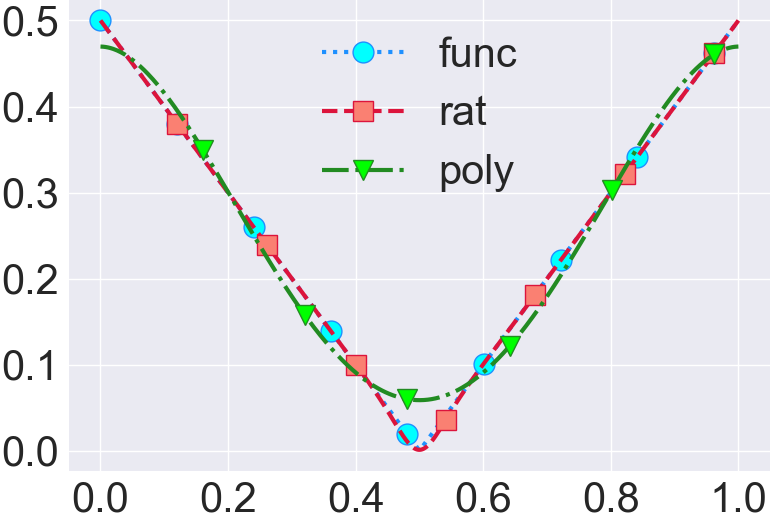}
    \end{subfigure}
    \begin{subfigure}
        \centering
        \includegraphics[width=0.2\textwidth]{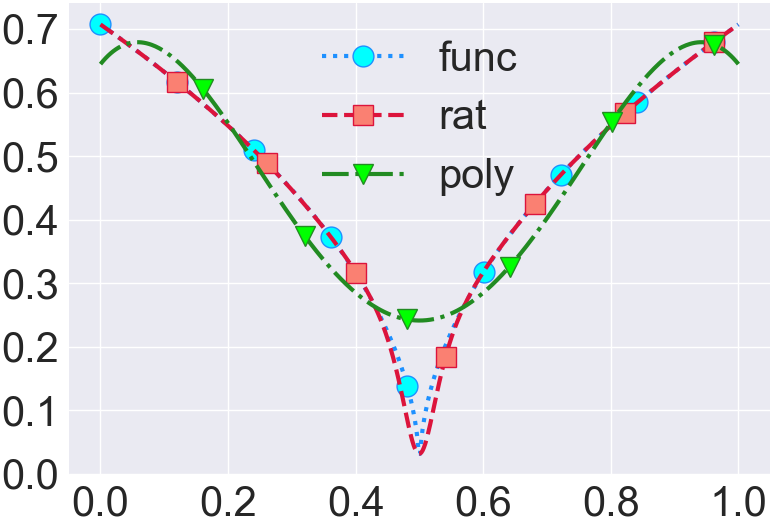}
    \end{subfigure}
    \begin{subfigure}
        \centering
        \includegraphics[width=0.2\textwidth]{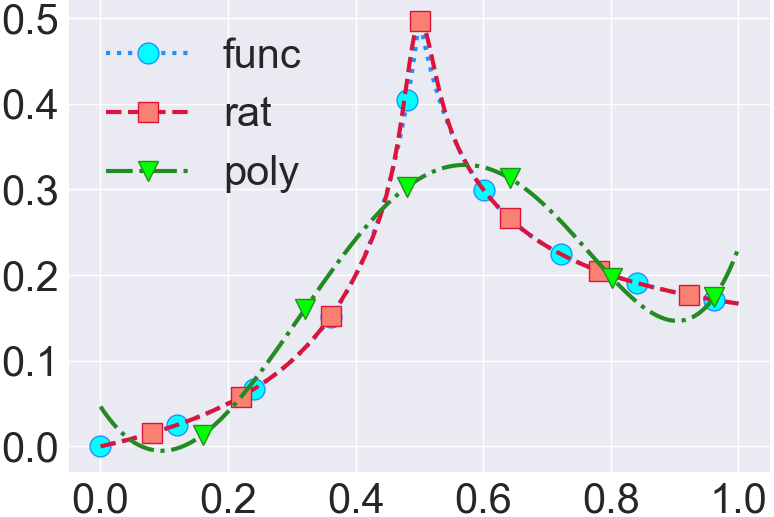}
    \end{subfigure}
    \begin{subfigure}
        \centering
        \includegraphics[width=0.2\textwidth]{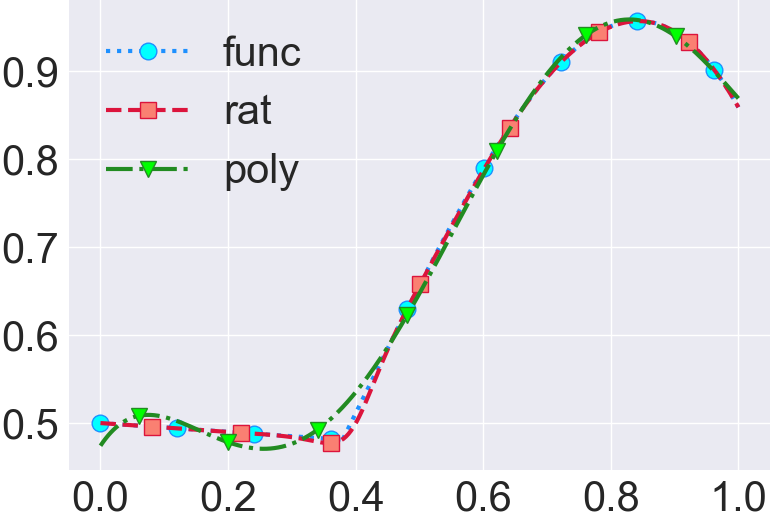}
    \end{subfigure}
    \caption{Rational (rat) and polynomial (poly) approximation for several functions with discontinuity (func). 
        From left to right: $\sqrt{|x-0.5|}$; $|x-0.5|$; $\frac{x}{10|x - 0.5|+1}$; $max(0.5, sin(x+x^{2}))-\frac{x}{20}$. Figures are from \cite{chen2018rational}.}
    \label{fig:fit_example}
    \vspace{-5pt}
\end{figure}

\vspace{5pt}
\noindent\textbf{Remark:}
When the function to approximate contains discontinuities, rational function has overwhelming advantage over the polynomials or linear functions. Figure \ref{fig:fit_example} illustrates the difference between rational and polynomial approximation. Theoretically, rational approximation only needs exponentially less orders than that of polynomial functions \cite{chen2018rational}.




\section{Theoretical Analysis} 
\label{sec:theory}

In terms of volume, spatial-based approaches outnumber spectrum-based methods in the literature \cite{bronstein2017geometric,zhang2018deep,zhou2018graph,wu2019comprehensive,abadal2020computing}, owing to the following reasons: (1) Spectral-based methods have a much higher computing overhead than spatial-based methods, and spectral methods are less intuitive than spatial methods. (2) Spatial-based approaches are convenient for model construction and scalability. 
However, from a spatial and spectral perspective, there is a trade-off; neither has a major advantage over the other. This section outlines numerous viewpoints that demonstrate the merits and limitations of such views.

\subsection{Uncertainty Principle: Global v.s. Local Perspectives}
\begin{figure*}[]
    \centering
    \includegraphics[width=0.9\linewidth]{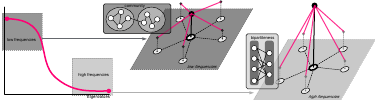}
    \caption{Connection between the Spatial and Spectral Perspective}
    \label{fig:global_and_local}
    \vspace{-10pt}
\end{figure*}

Spectral-based approaches decompose data into orthogonal frequency components and examine graph filtering from the spectral domain with a global perspective. Each frequency indicates a global basis: low-frequency components emphasize local weights with little variation, whereas high-frequency components are linked to significant variance in neighborhood. In other words, the Laplacian spectrum reflects topological properties: the first few eigenvalues are related with substantial community structure, whilst the last few eigenvalues indicate the graph's bipartiteness \cite{de2016role,de2014laplacian}. A typical low-pass filtering function for eigenvalues is shown on the left of Figure \ref{fig:global_and_local}, which raises small eigenvalues while decreasing adjusted values for large eigenvalues. Only low-frequency components are maintained in this scenario, and neighbors have little variance. 

Filtering patterns from the local neighborhood are characterized by spatial-based approaches. Most GNNs assume homophily among neighbors, so signals traversing across the neighborhood is smooth or with little variation, which is exactly the same concept as low-pass filtering. The relationship between low-pass filtering in the spectral domain (left) and its effects in the spatial domain (middle and right) is depicted in Figure \ref{fig:global_and_local} ~\cite{de2014laplacian}. 
It appears at first glance that global and local viewpoints differ greatly, but on closer inspection, they depict the same signal in very different ways: filtering in the spectral domain that does low-pass/high-pass filtering is analogous to learning which neighbors are similar/dissimilar.
While it is true that the two types of observations yield similar results, this does not entail that they are identical. It is impossible to know an unknown quantity's value with absolute confidence in quantum mechanics because of the Heisenberg's uncertainty principle \cite{folland1997uncertainty}. 
Specifically, 
\begin{equation}
    \Delta_{t}^{2} \Delta_{\omega}^{2} \geq \frac{1}{4},
\end{equation}where $\Delta_{t}$ and $\Delta_{\omega}$ denote time spread and frequency spread, respectively. 
Signal concentration can also be impacted by the concentration of time and frequency. 
A graph representing the trade-off between a signal's localization on a graph and in its spectral domain is created, which is influenced by the uncertainty principle of quantum mechanics \cite{agaskar2013spectral}. A lower bound on the product of the two spreads is obtained by quantifying the spreads in the vertex  and the spectral domain of a graph signal $x$. 
This result suggests that applying spatial-based GNNs results in accuracy loss in the spectral domain, while using spectral-based GNNs results in accuracy loss in the spatial domain.
In sum: 
\textbf{(1) Global and local perspectives are strong connected}: Global observation is the process of generalizing details in local settings, while localized understanding provides details of the global picture.
\textbf{(2) Global and local perspectives outperform each other in their own domains}: Clear local context tends to muddy the big picture, but a focus on global context diminishes the smaller picture.
\begin{table}[hpbt!]
\caption{Comparison between the Spatial (A-0) and Spectral (B-0) Methods}
\scalebox{0.8}{
\begin{tabular}{|c|c|c|c|c|}
\ChangeRT{2pt}
         & Methodology & Computation & Space Complexity & Stability   \\ \hline
Spectral & Global            & One-step    & High & Exact       \\ \hline
Spatial  & Local             & Iterative   & Low & Approximate \\ \ChangeRT{2pt}
\end{tabular}}
\label{tab:global_vs_local}
\end{table}
GNNs are able to expand the range of options with the addition of other preexisting works, which bridge the gap between global and local views or between spectral and spatial information, to improve the expressive potential of GNNs \cite{velivckovic2018deep,liang2020attributed,zhuang2018dual,zhang2019latentgnn,zhu2020beyond,Bo2020beyond}. According to the information above, we can state that no model can be flawless from a global or local perspective. It is only possible to have a proper trade-off between. As shown in Table \ref{tab:global_vs_local}, four aspects are compared:
\textbf{Methodology}: Spatial approaches describe local regions while working bottom-up, and identify global patterns using a graph frequency approach. On the other hand, spectral methods work top-down, beginning with a graph and ending with a global observation.
\textbf{Computation}: To use spatial approaches, one has to carry out a number of steps on their local region before convergence is achieved. With spectral approaches, you can get a critical component with a single-step computation.
\textbf{Space Complexity}: The high space complexity of spectral approaches is associated with the massive memory storage required to load the full graph. If memory is sufficient, the full graph can be covered by using spatial methods. However, for a smaller graph, you can choose to cover it using samplings such as sampling of regions or paths.
\textbf{Stability}: To create accurate, consistent results, spatial analysis methods need to apply iterative algorithms, therefore the outcomes will vary. Eigen-decomposition is a unique feature of spectral approaches if no same eigenvalues.

\noindent
\textbf{Guidance of Choosing Spatial and Spectral Methods.}
The user can choose between spatial or spectral GNNs depending on their previously described properties. 
\textbf{Distributed and Online Learning}: spatial method is easily converted to distributed learning \cite{lin2022comprehensive,shao2022distributed}, whereas spectral is difficult to transfer. Even if it is possible to approximate spectrum technique using a neural network model \cite{shaham2018spectralnet} and then use distributed learning on a neural network \cite{verbraeken2020survey}, new nodes and edges must be retrained from scratch. Alternatively, the spatial technique can effectively manage online learning with streaming data \cite{gao2022wide}. \textbf{Global View}: The spectral method may provide a global perspective that the spatial method lacks. In situations where the group form is not spherical, the spatial method may disregard this influence and continue to follow the circular shape as a prospective group \cite{he2022improved,azizyan2015efficient}. This can be remedied by employing clustering before to the spatial technique \cite{shi2022clustergnn}. This also renders spatial methods more locally interpretable and obscures their global perspective.

\subsection{Comparison between Linear, Polynomial and Rational Methods}\label{subsec:time_complexity}
Linear methods (A-1 and B-1) have a time complexity of $\mathcal{O}(N^2F)$ due to the matrix multiplication of $\A\X$. Accordingly, polynomial and rational method are analyzed in Table \ref{tab:analysis} where K is the order number. To compare their expressive power, the convergence rate on challenging jump signal is employed as a benchmark~\cite{chen2018rational} (a smooth signal cannot distinguish them). As shown in Table \ref{tab:analysis}, rational methods (A-3 and B-3) converge exponentially faster than linear methods (A-1/B-1), and polynomial methods (A-2/B-2) converge linearly faster than linear methods (A-1/B-1).
Therefore, there is a trade-off between the expressive power and computational efficiency. linear methods (A-1/B-1) have the best efficiency but only capture the linear relationship. Rational methods (A-3/B-3) consume the most considerable overhead but could tackle more challenging signals.
\begin{table}[hbpt!]
\centering
\caption{Comparison on Time Complexity and Expressive Power}
\scalebox{0.8}{
\begin{tabular}{|l|l|l|l|}
\ChangeRT{2pt}
             & Linear (A-1, B-1) & Polynomial (A-2, B-2) & Rational (A-3, B-3) \\ \ChangeRT{1pt}
Time         &  $\mathcal{O}(N^2F)$ & $\mathcal{O}(N^{K+1}F)$ & $\mathcal{O}(N^{K+1}F+N^3)$     \\ \hline
Expressivity &  $\mathcal{O}(1)$    & $\mathcal{O}(1/K)$ & $\mathcal{O}(\exp^{-\sqrt{K}})$          \\ \ChangeRT{2pt}
\end{tabular}}
\label{tab:analysis}
    \vspace{-1pt}
\end{table}


Experiments are undertaken to highlight our theoretical analysis of spatial and spectral approaches by comparing the differences between the three underlying groups. 
We chose one typical technique for linear filter \cite{kipf2016semi}, polynomial filter \cite{defferrard2016convolutional}, and rational filter \cite{bianchi2019graph}. Note that we only distinguish them in the function of $\A$ or $\LBD$, keeping all the other configurations the same. The dataset includes representative homophily and heterophily datasets \cite{acm_neurips22,asgc_neurips22,ugcn_neurips21,dmp_neurips21,pgnn_icml22,glognn_icml22}.
The evaluation code is released \footnote{\url{https://github.com/aquastar/csur_bridge_spectral_spatial_gnn_survey}}. The implementation is based on the official Pytorch Geometric \cite{pygdoc}. Each model on each dataset is evaluated 50 times, and the results are averaged.

\begin{figure*}[htpb!]
    \centering
    \begin{subfigure}
        \centering
        \includegraphics[width=0.45\linewidth]{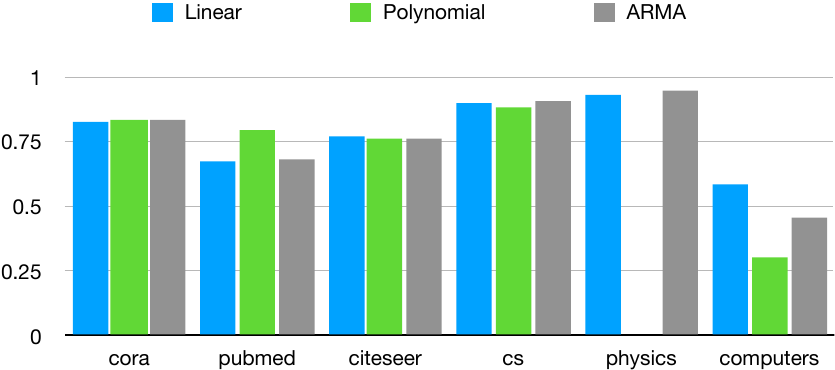}
    \end{subfigure}%
    ~ 
    \begin{subfigure}
        \centering
        \includegraphics[width=0.45\linewidth]{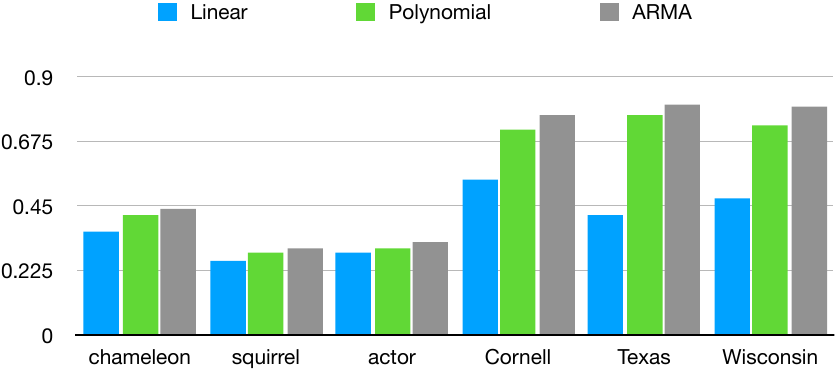}
    \end{subfigure}
    \caption{Performance Comparison. \textbf{Left}: homophily dataset; \textbf{Right}: heterophily dataset}
    \label{fig:acc_comp}
\end{figure*}

\begin{figure*}[t!]
    \centering
    \begin{subfigure}
        \centering
        \includegraphics[width=0.45\linewidth]{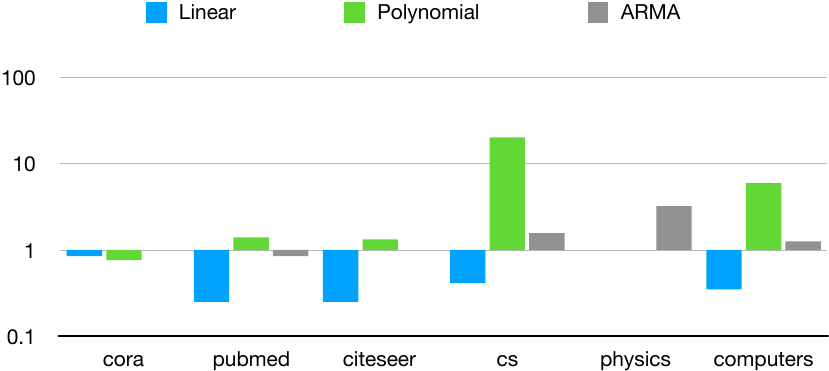}
    \end{subfigure}%
    ~ 
    \begin{subfigure}
        \centering
        \includegraphics[width=0.45\linewidth]{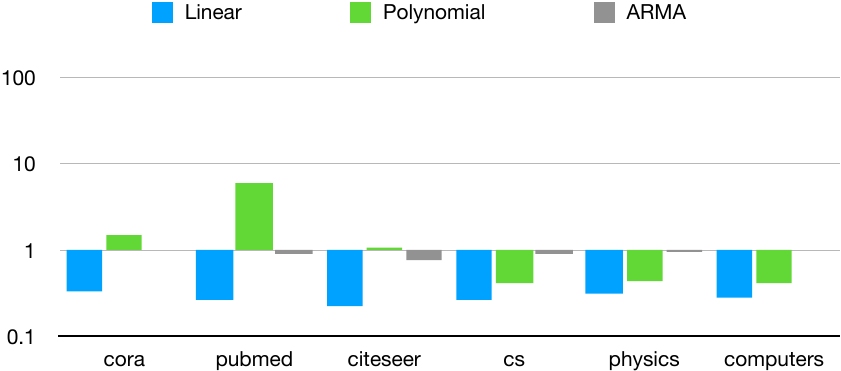}
    \end{subfigure}
    \caption{Runtime Comparison. \textbf{Left}: homophily dataset; \textbf{Right}: heterophily dataset. (Y-axis is log-based)}
    \label{fig:time_comp}
\end{figure*}

As shown in Figure \ref{fig:acc_comp} (Left), there is no significant difference in classification accuracy in the homophily dataset between the three models, with the exception of ChebNet, which performs marginally better in \textit{PubMed} and encounters an out-of-memory error in \textit{physics} dataset. ChebNet exhibits inferior accuracy on the \textit{Computers} dataset, whereas GCN is significantly superior. Figure \ref{fig:acc_comp}  (Right) illustrates that ARMA consistently outperforms the others when performing the same task on the heterophily dataset, while ChebNet consistently outperforms GCN. This demonstrates the benefits of the advanced filter function, as theoretically analyzed in Section \ref{subsec:btwn_spa_spec}.
In terms of runtime, as shown in Figure \ref{fig:time_comp}, ChebNet is often beyond $\log 1$, while GCN never goes beyond $\log 1$. Rational method involve more matrix operation as ChebNet does, so rational method sometimes goes beyond $\log 1$,  but due to optimization in implementation (iterative algorithm), which makee it a little faster than ChebNet, but still consistently slower than GCN.
In terms of runtime, as shown in Figure \ref{fig:time_comp}, ChebNet frequently exceeds $\log 1$, whereas GCN never does. Rational method involves more matrix operations than ChebNet, so it sometimes exceeds $\log 1$. However, due to optimization in implementation (iterative algorithm), the rational method is slightly faster than ChebNet but consistently slower than GCN. This verifies our analysis in Section \ref{subsec:time_complexity}.

\vspace{-5pt}
\section{Exemplify the Proposed Framework}
\label{sec:example}
Over-smoothing and large-scale difficulties are two of the most difficult issues for existing GNNs, and numerous recent publications have proposed various approaches to address them. We'll show in this section that all of the enhancements are still covered by our framework.

\subsection{Sampling Point of View}\label{sec:sampling}
The sampling mechanism is used as a spatial method for managing large graphs. Subgraph sampling and random walk are popular approaches.

\subsubsection{Subgraph Sampling}
For an early work, \textit{GraphSAGE}~\cite{hamilton2017inductive} applies uniform node sampling for each batch, which is equivalent to subgraph sampling. The likelihood of transfer therefore follows random normalization (i.e., $\An=\D^{-1}\A$), which makes it part of Linear Aggregation (A-1).
In the majority of follow-up works, the same methodology is used~\cite{zeng2019graphsaint}: (1) build a local graph convolution for the input graph. (2) sample nodes in each layer, and (3) optimize parameters in graph convolution. Steps (2) and (3) proceed iteratively to update the weights via stochastic gradient descent
 \cite{zeng2019graphsaint,chen2018fastgcn,chen2018stochastic,huang2018adaptive,gao2018large,bns_ijcai21}.

To avoid the recursive neighborhood expansion,  FastGCN~\cite{chen2018fastgcn} treats graph convolutions as integral transformation of embedding functions and proposes Monte Carlo approach to estimate the integral. FastGCN employs importance sampling independently for each layer and reduce variance cutting down the number of sampling nodes to constant size for all layers, exponentially shrinking the computational cost. 
FastGCN is proved to be importance sampling, which is better than uniform sampling, but still suffers from unstable learning when no neighbors is selected for one node and activation is zero.
To avoid taxing calculation of activation, Stochastic GCN \cite{chen2018stochastic} further uses the historical activation in the previous layer to avoid redundant re-evaluation. 
With adaptive sampling, nodes on subsequent layers are sampled in order to speed up GraphSAGE and FastGCN \cite{huang2018adaptive}. Learnable graph convolutional layer (LGCL) \cite{gao2018large} selects a fixed number of neighboring nodes for each feature based on value ranking, and transform graph into 1-D data which is compatible with normal convolution networks. Similarly, A scalable GCN samples a fixed number of nodes, with different sampling policy called frontier sampling (FS). FS maintains a constant size frontier set consisting of several vertices which is randomly popped out with a degree based probability distribution \cite{zeng2019accurate}. Cluster-GCN~\cite{chiang2019cluster} samples a community of nodes determined by a graph clustering algorithm, and compute the graph convolution within each community.

\subsubsection{Random Walk}
To derive node-level representations with word2vec~\cite{mikolov2013linguistic,mikolov2013efficient,mikolov2013distributed}, various random walk algorithms are proposed \cite{gpnn_aaai22,grover2016node2vec,perozzi2014deepwalk,tang2015line,wang2016structural,zeng2019graphsaint}. Paths are viewed as complete sentences, and nodes are viewed as individual words. Transition probability among nodes approximates to a random walk normalized adjacency matrix if enough random walks or uniform sampling have been performed on the paths.

As analyzed in previous section, \textit{DeepWalk}~\cite{perozzi2014deepwalk} draws a number of random paths from the graph, which makes the transfer probability of random walk is, i.e., $\An=\D^{-1}\A$.
Let the window size of skip-gram be $2t+1$ and the index of current node is $t+1$. Therefore, the updated representation is as $\Z=\frac{1}{t+1}\Pp(\An)\X$ (Equation \ref{eq:a2_deepwalk}).
One popular word2vec configuration, i.e., skip-gram with Negative Sampling (SGNS), assumes a corpus of words w and their context c. Following the work by Levy and Goldberg \cite{levy2014neural}, SGNS is implicitly factorizing:
\begin{equation*}\small
    \log \left(\frac{|(w, c)|\cdot|\mathcal{D}|}{|w|\cdot|c|}\right)-\log b =\boldsymbol{A} \boldsymbol{B}^{\top},
\end{equation*}where $\boldsymbol{A}$ and $\boldsymbol{B}$ denote matrix of current node and its neighbors, respectively.
$|(w,c)|$, $|w|$, $|c|$ and $\mathcal{D}$ denote the number of times word-context pair $(w,c)$, word $w$, context $c$ and corpus size, respectively; $b$ is the number of negative samples. Accordingly, \cite{qiu2018network} derived a exact format as:
\begin{equation*}\small
    \log \left(\Pp(\An)\right)-\log (b) = \log \left(\frac{|E|}{T}\left(\sum_{r=1}^{T} (\D^{-1}\A)^{r}\right) \D^{-1}\right)-\log (b),
\end{equation*}where $|E|$ and $T$ represents edge number and step size, respectively. Therefore, the target matrix to decompose is still a polynomials of $\A$.
\textit{Node2Vec}~\cite{grover2016node2vec} defines a 2nd order random walk to control the balance between Breath First Search (BFS) and Depth First Search (DFS).  
Assuming the random walk is sufficiently sampled, Node2Vec's second order can be rewritten to decompose matrix \cite{qiu2018network}:
\begin{equation*}\small
    \log \left(\Pp(\An)\right)-\log (b) =\log \left(\frac{\frac{1}{2 T} \sum_{r=1}^{T}\left(\sum_{u} X_{w, u} \underline{\An}_{c, w, u}^{r}+\sum_{u} X_{c, u} \underline{\An}_{w, c, u}^{r}\right)}{\left(\sum_{u} X_{w, u}\right)\left(\sum_{u} X_{c, u}\right)}\right)-\log b,
\end{equation*}where Node2Vec is demonstrated to be polynomial methods.
\textit{LINE}~\cite{tang2015line} and \textit{SDNE}~\cite{wang2016structural} learn the node representations within the first- and second-order neighbors, which can be treated as unconstrained version of Node2Vec:
\begin{equation*}\small
    \log \left(\Pp(\An)\right)-\log (b)  = \log \left(|E|\An\right)-\log b.
\end{equation*}
GraphSAINT~\cite{zeng2019graphsaint} employs multiple sampling polices but the best one is random walk. \textit{PinSAGE}~\cite{ying2018graph} improves the efficiency of GraphSAGE~\cite{hamilton2017inductive} by taking the top several neighbors with highest normalized visit counts.

\vspace{2pt}
\noindent\textbf{Remark}: 
Sampling methods, as a spatial methodology, seek both variance reduction and efficiency. 
Subgraph and random walk are equal with enough samples since they traverse the entire network with the transition probability associated with graph connection. 
A-3 and B-3 do not, however, include any sample methods, mainly due to their higher computational complexity. Space complexity may be improved when sampling methods are used, but there is no guarantee that the time complexity will be much reduced because of the enormous number of steps may be needed before convergence.

\vspace{-5pt}
\subsection{Over-smoothing Point of View}\label{sec:over_smooth}
We carve out two conditions under which neighborhood aggregation is not helpful: (1) when a node’s neighbors are highly dissimilar and (2) when a node’s embedding is already similar to that of its neighbors. 

Most GNNs perform poorly when stacking many layers, which is called the over-smoothing issue.
Many related works aiming to solve the over-smoothing issue ~\cite{yang2020revisiting,zhu2020beyond,bo2021beyond,li2019deepgcns,zhao2019pairnorm,xu2018representation,oono2019graph,rong2019dropedge,li2020deepergcn,huang2020tackling,chen2020simple,liu2020towards,zhou2020deeper,chen2020measuring,min2020scattering} can be reduced to one category of our proposed framework.
H2GCN~\cite{zhu2020beyond} proposed a method that combines direct neighbors with higher-order, which is equivalent to Polynomial Aggregation (A-2). 
Deep GCN~\cite{li2019deepgcns,yang2020revisiting,li2020deepergcn} developed a model with a residue module, dense connection, and dilated aggregation, which learns the weights of all different orders of neighbors. This is equivalent to Polynomial Aggregation (A-2). 
GPR \cite{chien2020adaptive} generalize PageRank and found the equivalence of GPR and Polynomial Approximation (B-2).
JKNet~\cite{xu2018representation} also follows the same residue methodology as Deep GCN.
DAGNN~\cite{liu2020towards} stacks multiple layers which uses different orders of propagation with the learnable weights, which makes it belong to Polynomial Aggregation (A-2).
PairNorm~\cite{zhao2019pairnorm} presents a two-step method that includes centering and re-scaling, which mitigates the over-smoothing from graph convolution. Therefore, PairNorm is equivalent to Rational Propagation (A-3), since re-scaling is similar to do propagation and restart at the same time.
\cite{bo2021beyond} design an adaptive method to dynamically adjust the weights between low-frequency and high-frequency components, resulting in two peaks in the spectral domain. This could also be modeled by Rational Aggregation (A-3) with its accuracy in jump signals.
DropEdge~\cite{rong2019dropedge,huang2020tackling} randomly drops a certain number of edges to avoid over-smoothing, which can be categorized as Rational Aggregation (A-3) since dropping edge prevents the propagation and thereby provides a probability of keeping the original values of nodes. 
GCNII~\cite{chen2020simple} applies \textit{initial residual} which combines the smoothed representation with an initial residual connection to the first layer, and \textit{identity mapping}, which adds an identity matrix to the weight matrix. \textit{Initial residual} is a trick that PPAP~\cite{klicpera2018predict} uses, which enables itself to retard the over-smoothing by keeping partial previous representations. \textit{Identity mapping} further remains one original representation to slow down the spreading of over-smoothing propagation.

\vspace{2pt}
\noindent\textbf{Remark}: 
A-1 lacks state-of-the-art methods, implying it is vulnerable to over-smoothing. Applying A-1 many times with learnable weights for different ordering equates to A-2. So A-2 might balance low (raw representations) and high orders (smoothed representations). However, too many A-1 operations may result in over-smoothing. So, for A-2, precise order configuration is required. No matter the number of orders or layers, A-3 reserves a proportion of the final representation as raw representation, making itself robust to over-smoothing.

\section{Limit, Open Challenges and Conclusion}
\label{sec:open}




In this paper, we present a unifying paradigm for comprehending GNNs created under various processes. Our study shows that the subcategories are closely related via generalization and specialization links within their domains, as well as equivalence ties across domains. We show the framework's generalization power by reformulating existing GNNs models. 
As introduced in the sections above, spatial methods are designed by various ideas, and they can be interpreted well by the unified spectral theory. Therefore, we will discuss the potential that spectral theory, as a unfied theoretical framework, may also extend to emerging directions.

Despite the an increasing number of emerging GNNs models \cite{errica2019fair} made in recent years, spectral methods so far has been intensively studied in node-level graph convolution only, leaving the other graph learning problem uncovered. 
In recent years, graph learning has been successfully  extended to various tasks such as (sub)graph-level tasks, combinatorial optimization, explainability, domain application (brain, PDE solver, circuit, molecule, protein), generative graph, graph transformer, contrast learning, heterogeneous graph  \cite{guo2019learning,baldassarre2019explainability,ying2019gnn,loukas2019graph,oonograph}. However, a unified framework is lacking, which is due to the underlying theories from spectral graph theory and graph signal processing applied in node-level graph convolution application by the first few pioneer work \cite{hammond2011wavelets,defferrard2016convolutional,kipf2016semi}, but little attention has been paid to the theoretical study on the emerging topics rather than node-level convolution. Therefore, most new topics in graph machine learning is based on intuitive design or isolated theory, lacking a unified framework to compare and understand these separate learning models.
In this section, we will list the possibility that all the selected topics can be looked at through a unified framework.

\vspace{5pt}\noindent\textbf{Theoretical Understanding -}The vast majority of recent research has examined spatial and spectral approaches individually, and even from a theoretical standpoint. \cite{xu2018how,jacobiconv_icml22,kenlay2021interpretable}. However, it is unclear how these two distinct interpretations can be connected. This survey demonstrates that the theoretical advantage of rational function over the others may be proved, however it is currently unclear how the learning approach can be optimally structured to execute this advantage. Using the partial differential equation \cite{strauss2007partial}, in which the diffusion equation and wave equation are analogous to polynomial and rational filters, provides an alternate viewpoint to integrate spatial and spectral views.

\vspace{2pt}\noindent\textbf{Directed Graph -}
Most contemporary work, especially spectral methods, only handle undirected graphs, due to symmetric graph matrix is readily available with off-the-shelf techniques. Many related works integrate or sum up asymmetric adjacency matrices from bi-direction into symmetric matrix, which avoids decomposing asymmetric matrix directly.
It is possible to decompose asymmetric matrix by some techniques such as the Jordan norm \cite{kaagstrom1980algorithm}, asymmetric matrices can be used to express properties such as graph and filter complexity. Directed graphs and their decomposition can also be achievable in a certain type of geometric space called a graph manifold. As shown in section 2.4 of \cite{levie2021transferability}, spectral filters can be defined on directed graphs represented by non-symmetric adjacency matrices with Hermitian transpose.

\vspace{2pt}\noindent\textbf{Dynamic Graph -} 
Existing work model with GNNs and RNNs is built using graph convolutional networks and recurrent neural networks, which lack transparency. 
Due to the limited expressive power of RNNs, the task is constrained to prediction, and it is unable to perform long-term sequence processing well. 
Graph dynamics has a large number of valuable tasks, such as inferring the structure of a graph, constructing a joint dynamic of structure and attributes, and exploring the connection between structure and mass flow. 
It is possible to use graph spectra to detect the patterns in dynamics of graph \cite{malik2012dyn,silva2018spec,bronski2014spectral}.
Due to the challenge of the long-dependency of a path, the spectral method can also provide a potential way to model trajectory prediction, as the spectra of trajectory implicitly reveal the relationship with the whole graph \cite{cao2021spectral}.

\vspace{2pt}\noindent\textbf{Higher-Order Interaction and Combinatorial Optimization -} Most current work falls into first-order relationship at node-level. For example, the most spatial method is to learn the relationship between the current node and neighbors, but higher-order interaction is either ignored or implicitly included. Existing explainable learning also focuses on the neighbors' identification \cite{yuan2022explainability}. Also, the existing work pays the most attention to neighbors, but remote connection or higher-order relationship with the other nodes receives little attention.
Hypergraphs provide a possibility to model the combinatorial effect \cite{bai2021hypergraph,feng2019hypergraph,hgcrnn_kdd20}
Hodge Laplacian \cite{ribando2022graph,lim2020hodge} and simplicial complex \cite{bianconi2021higher,benson2018simplicial,battiston2020networks}  provide more theoretical tools for modeling this combinatorial effect.


\vspace{2pt}\noindent\textbf{Multilayer Network -}
In numerous realistic biological and engineering systems, however, the units can be interconnected and interdependent via multiple interdependent and heterogeneous networks. Failure of interdependent nodes between linked networks may result in cascading failures inside and across the networks.
Similar interactions exist in many cyber-physical systems, where the spread of misinformation about infectious diseases via social media can result in risky daily plans at the group level, resulting in an epidemic outbreak.
Such interconnected networks can be represented by \textit{multilayer networks} that produce new degrees of freedom via coupling interactions. Such ``new physics'' is prevalent in multilayer systems, but they are still poorly understood \cite{de2016physics,bianconi2018multilayer,aleta2018multilayer,dong2021briefing}. 
Graph neural network research on this crucial area is scarce \cite{grassia2021mgnn}. One popular technique to generalize principles from monolayer networks to multilayer networks is to ``flatten'' adjacency tensors into matrices (called ``supra-adjacency matrice''), and spectral theory is available and worth researching.

\bibliographystyle{ACM-Reference-Format}
\bibliography{z}

\newpage
\appendix
\section{B-1}
\subsubsection{\textbf{Graph Convolutional Network (GCN)}} Rewriting GCN \cite{kipf2016semi} in spectral domain, we have:
\begin{equation}\small
\Z = \An\X=\Dsym(\A+\I)\Dsym\X=\Dsym(\D-\Ll+\I)\Dsym\X=(\I-\Dsym\Ll\Dsym+\I)\X=\UU(2-\LBD)\UT\X,
\label{eq:a1_gcn2}
\end{equation}
where $\An={\D}^{-\frac{1}{2}}(\A+\I){\D}^{-\frac{1}{2}}$ is renormalization of $\An$. 
Therefore, the frequency response function is $\g(\LBD) = 1-\LBD$ which is a low-pass filter, i.e., smaller eigenvalues which correspond to low frequency components are assigned with a larger value.

\subsubsection{\textbf{GraphSAGE}} Considering the MEAN aggregation as example, we can rewrite GraphSAGE \cite{hamilton2017inductive} in matrix form:
\begin{equation}\small
\Z = \Dsym(\I+\A)\Dsym\X=(\I+\An)\X=(2\I-\Ln)\X=\UU (2-\LBD) \UT\X.
\label{eq:a1_gcn1}
\end{equation} 

Hence, the frequency response function is $\g(\LBD) = 2-\LBD$ which is a low-pass filtering. Note that GraphSAGE's normalization is different from GCN, which utilizes the renormalization trick.
\begin{multicols}{2}
\begin{equation}
      (I + D^{-1} A)\X 
\label{eq:gcn_spatial}
\end{equation}
\break
\begin{equation}
    \UU \big( 1 + \Lambda_{P} \big) \UT \X
\label{eq:gcn_spectral}
\end{equation}
\end{multicols}

where $P = D^{-1} A$

\subsubsection{\textbf{Graph Isomorphism Network (GIN)}} 

Multi-Layer neural network is capable of fit the scale (i.e., normalization) \cite{kipf2016semi}, so GIN \cite{xu2018how} can be rewritten as:
\begin{equation}\small
    \Z = \Dsym [(1+\epsilon)\I + \A]\Dsym\X=\Dsym [(2+\epsilon)\I - \Ln]\Dsym\X= \UU(2+\epsilon - \LBD)\UT\X.
\end{equation} 

GIN can be seen as a generalization of GCN or GraphSAGE without normalized adjacency matrix $\A$. The frequency response function is $\g(\LBD) = 2+\epsilon - \LBD$ which is low-pass.

\section{B-2}

\subsubsection{\textbf{ChebNet}} As analyzed in Equation \ref{eq:a2_chebnet}, ChebNet \cite{hammond2011wavelets} can be written as:
\begin{equation}\small
    \sum_{k=0}^{K-1} \theta_{k} T_{k}(\tilde{\Ll}) \X= (\tilde{\theta_{0}}\I + \tilde{\theta_{1}}\tilde{\Ll} + \tilde{\theta_{2}}\tilde{\Ll}^{2} +... ) \X,
\end{equation}
where $T_{k}(\cdot)$ is the Chebyshev polynomial and $\theta_{k}$ is the Chebyshev coefficient. $\tilde{\theta}$ is the coefficient after expansion and reorganization. Therefore, we can rewrite it as:
\begin{equation}\small
        \sum_{k=0}^{K-1} \theta_{k} T_{k}(\tilde{\Ll}) \X= \UU (\tilde{\theta_{0}}\cdot 1 + \tilde{\theta_{1}}\LBD + \tilde{\theta_{2}}{\LBD}^{2} +... )\UT \X,
\end{equation}where spectral response function is $\g(\LBD)=\tilde{\theta_{0}}\cdot 1 + \tilde{\theta_{1}}\LBD + \tilde{\theta_{2}}{\LBD}^{2} +... =\Pp(\LBD)$.

\subsubsection{\textbf{DeepWalk}} Starting from Equation \ref{eq:a2_deepwalk}, DeepWalk \cite{perozzi2014deepwalk} can be rewritten as:
{\small
\begin{align*}
    \Z &= \frac{1}{t+1}(\I + \An + \An^{2} + ... + \An^{t})\X&\\
    &= \frac{1}{t+1}(\I + (\I-\Ln) + (\I-\Ln)^{2} + ... + (\I-\Ln)^{t})\X &\\
    &=\frac{1}{t+1}(2\I  + (-1-2-3-...)\Ln + (1+3+6+...)\Ln^{2} + 
    ... ((-1)^{t-1}+\left(\begin{array}{l}{1}\\{t}\end{array}\right) (-1)^{t-1}) \Ln^{t-1}+ (-1)^{t}\Ln^{t})\X &\\
    &=(\theta_{0}\I  + \theta_{1}\Ln +\theta_{2}\Ln^{2} + ... + \theta_{t}\Ln^{t})\X &\\
    &=\UU(\theta_{0} + \theta_{1}\LBD +\theta_{2}\LBD^{2} + ... + \theta_{t}\LBD^{t})\UT\X&\\
    &=\UU\Pp_{k=r}(\LBD)\UT\X,
\end{align*}}where $\g(\LBD)=\theta_{0} + \theta_{1}\LBD +\theta_{2}\LBD^{2} + ... + \theta_{t}\LBD^{t}$, and all parameters $\theta_{i}$ are determined by the predefined step size t.

\subsubsection{\textbf{Scalable Inception Graph Neural Networks (SIGN)}}
Substituting $\An = \I -\Ln$ in Equation \ref{eq:a2_sign}, it can be rewritten as:
\begin{equation}\small
    \hat{\Z} = \sum_{r} \omega_{r} \widehat{(\I-\Ln)^{r}\X} =\UU \widehat{\Pp_{k=r}(\LBD)\X}\UT.
\end{equation}

\subsubsection{\textbf{Graph diffusion convolution (GDC)}}
Substituting $\An = \I -\Ln$, general case in Equation \ref{eq:a2_gdc} can be written as:
\begin{equation}\small
    \Z=\sum_{k=0}^{\infty} \theta_{k} (\I-\Ln) = \UU\sum_{k=0}^{\infty} \theta_{k} (1-\LBD)^k \UT=\UU\Pp(\LBD)\X\UT.
\end{equation}

\subsubsection{\textbf{Diffusion convolutional neural networks (DCNN)}} As analyzed in Equation \ref{eq:a2_dcnn}, DCNN \cite{atwood2016diffusion} can be transformed with $\An = \I -\Ln$ as:
\begin{equation}\small
  \Z=\Pp(\I-\Ln)\X = \UU\Pp(\LBD)\X\UT,
\end{equation} which is equivalent to ChebNet, and parameters $\theta_{i}$ are learnable.

\subsubsection{\textbf{Node2Vec}}
Node2Vec \cite{grover2016node2vec} can be rewritten in matrix form as Equation \ref{eq:node2vec_mat}. Then it can be transformed and reorganized after substituting $\An=\I-\Ln$:
\begin{equation}
    \Z =[(1+\frac{1}{p})\I -(1+\frac{1}{q})\Ln +\frac{1}{q}\Ln^{2}]\X =\UU [(1+\frac{1}{p})-(1+\frac{1}{q})\LBD +\frac{1}{q}\LBD ^{2}]\UT\X.
\end{equation} 

Therefore, Node2Vec's frequency response function is: 
\begin{equation}\small
    \g(\LBD)=(1+\frac{1}{p})-(1+\frac{1}{q})\LBD +\frac{1}{q}\LBD ^{2},
\end{equation}
which integrates a second order function of $\LBD$ with predefined parameters, i.e., $p$ and $q$.

\subsubsection{\textbf{LINE/SDNE} }
As described in Equation \ref{eq:a2_line_sdne}, LINE \cite{lin2015learning} and SDNE \cite{wang2016structural} can be rewritten as:
\begin{equation}\small
    \Z = \An\X+\alpha\An^2\X= (\I-\Ln)\X+\alpha(\I-\Ln)^2\X=\UU[(\I-\LBD)+\alpha(1-\LBD)^2]\X\UT=\UU\Pp_{k=2}(\LBD)\UT\X,
\end{equation}where response function $\g(\LBD)=\LBD+\alpha\LBD^2$ is a polynomial function with order 2.

\subsubsection{\textbf{Simple Graph Convolution (SGC)}}
As analyzed in Equation \ref{eq:a2_sgc}, SGC can be transformed as:
\begin{align*}
\Z  = (\I-\Ln)^{k}\X &=[\left(\begin{array}{l}{k} \\ {0}\end{array}\right) \I +\left(\begin{array}{c}{k} \\ {1}\end{array}\right) \Ln^{1}+\left(\begin{array}{c}{k} \\ {2}\end{array}\right)  \Ln^{2}+\cdots+ \Ln^{k}]\X&\\
    &=\UU[\left(\begin{array}{l}{k} \\ {0}\end{array}\right) +\left(\begin{array}{c}{k} \\ {1}\end{array}\right) \LBD^{1}+\left(\begin{array}{c}{k} \\ {2}\end{array}\right)  \LBD^{2}+\cdots+ \LBD^{k}]\UT\X,&
\end{align*}where spectral response function is a polynomial function of order k:
\begin{equation*}\small
    \g(\LBD)=\left(\begin{array}{l}{k} \\ {0}\end{array}\right) +\left(\begin{array}{c}{k} \\ {1}\end{array}\right) \LBD^{1}+\left(\begin{array}{c}{k} \\ {2}\end{array}\right)  \LBD^{2}+\cdots+ \LBD^{k}.
\end{equation*}

\subsubsection{\textbf{Improved GCN (IGCN)}}
By stacking multiple layers IGCN \cite{Li_2019_CVPR} is proposed as:
\begin{equation}\small
    \Z =\Ln^{k} \X = \UU\LBD^{k}\UT \X,
\end{equation}where the spectral response function is a polynomial function with order k.

\section{B-3}
\subsubsection{\textbf{Auto-Regressive filter}}
Label propagation (LP) \cite{zhu2003semi,zhou2004learning,bengio200611} is a prevail methodology for graph-based learning. The objective of LP is two-fold: one is to extract embeddings that matches with the label, the other is to be similar with neighboring vertices. Label can be treated as part of node attributes, so we can have:
\begin{equation}\small
    \Z =(\I +\alpha \Ln )^{-1} \X = \UU\frac{1}{1+\alpha (1-\LBD)}\UT \X.
\end{equation}

\subsubsection{\textbf{PPNP}}
Personalized PageRank (PPNP) \cite{klicpera2018predict} can obtain node's representation via teleport (restart) probability $\alpha$ which indicates the ratio of keeping the original representation:
\begin{equation}\small
    \Z= \frac{\alpha }{\I-(1-\alpha )(\I-\Ln )} \X =\UU\frac{\alpha}{\alpha+(1-\alpha) \LBD}\UT \X,
\label{eq:ppnp_b3}
\end{equation} where $\An=\Drw\A$ is random-walk normalized adjacency matrix with self-loop. Equation \ref{eq:ppnp_b3} is with a rational function whose numerator is a constant.


\subsubsection{\textbf{ARMA filter}} Substituting $\An=\I-\Ln$, Equation \ref{eq:a3_arma} can be rewritten as:
\begin{equation}\small
    \Z=\frac{b}{\I-a(\I-\Ln)}\X=\UU\frac{b}{(1-a)+a\LBD}\UT\X.
\end{equation}

Note that ARMA filter is an unnormalized version of PPNP. When a+b=1, ARMA filter becomes PPNP. Therefore, ARMA filter is more generalized than PPNP due to its unnormalization.

\subsubsection{\textbf{ParWalks}}\cite{li2018deeper,wu2012learning}
Decomposing graph Lapacian, ParWalks can be written as:

\begin{equation}\small
    \Z=\UU\frac{\beta}{\beta + \LBD}\X\UT,
\end{equation}
when setting $\beta=\frac{\alpha}{1-\alpha}$, it becomes PPNP:
\begin{equation}\small
    \Z=\UU\frac{\frac{\alpha}{1-\alpha}}{\frac{\alpha}{1-\alpha} +\LBD}\X\UT=\UU\frac{\alpha}{\alpha +(1-\alpha)\LBD}\X\UT.
\end{equation}

\subsubsection{\textbf{RationalNet}}
Substituting $\An=\I-\Ln$, Equation \ref{eq:a3_mat} can be transform to Equation \ref{eq:b3}. The frequency response function is a generalized rational function.

\end{document}